\documentclass[sigconf]{acmart}
\usepackage{epsfig}
\usepackage{graphicx}
\usepackage{amsmath,epsfig}
\usepackage{algorithmic}
\usepackage{graphicx}
\usepackage{textcomp}
\usepackage{xcolor}
\usepackage{rotating}
\usepackage{multirow}
\usepackage{booktabs}
\usepackage{colortbl}
\usepackage{array}
\usepackage{overpic}
\usepackage{float}

\hypersetup{
  urlcolor=red
}

\renewcommand\footnotetextcopyrightpermission[1]{} % removes footnote with conference information in first column

\def\ie{\emph{~i.e.,~}}
\def\eg{\emph{,~e.g.,~}}

\def\etal{{\em ~et al.~}}
\def\ourmodel {\emph{DFM-Net}}
\definecolor{mygray}{gray}{.92}

\AtBeginDocument{%
  \providecommand\BibTeX{{%
    \normalfont B\kern-0.5em{\scshape i\kern-0.25em b}\kern-0.8em\TeX}}}

\setcopyright{acmcopyright}
\copyrightyear{2018}
\acmYear{2018}
\acmDOI{10.1145/1122445.1122456}

\acmConference[Conference’21]{ACM Conference}{ October 2021}{Chengdu, China}
\acmBooktitle{Depth Quality-Inspired Feature Manipulation for Efficient RGB-D Salient Object Detection}
\acmPrice{15.00}
\acmISBN{978-1-4503-XXXX-X/18/06}

\acmSubmissionID{433}
\begin{document}

%%
%% The "title" command has an optional parameter,
%% allowing the author to define a "short title" to be used in page headers.
\title{Depth Quality-Inspired Feature Manipulation \\for Efficient RGB-D Salient Object Detection}

%%
%% The "author" command and its associated commands are used to define
%% the authors and their affiliations.
%% Of note is the shared affiliation of the first two authors, and the
%% "authornote" and "authornotemark" commands
%% used to denote shared contribution to the research.
\author{Wenbo Zhang}
%\authornote{Both authors contributed equally to this research.}
%\email{trovato@corporation.com}
%\orcid{1234-5678-9012}
%\author{G.K.M. Tobin}
%\authornotemark[1]

\affiliation{
\institution{College of CS, Sichuan University}
\country{}
}
\email{zhangwenbo@stu.scu.edu.cn}

\author{Ge-Peng Ji}
\affiliation{
\institution{Inception Institute of AI}
\country{}
}
\email{gepengai.ji@gmail.com}

\author{Zhuo Wang}
\affiliation{
\country{}
\institution{College of CS, Sichuan University}
}
\email{ imwangzhuo@stu.scu.edu.cn}

\author{Keren Fu*}
\affiliation{
\country{}
\institution{College of CS, Sichuan University}
}
\email{fkrsuper@scu.edu.cn}

\author{Qijun Zhao}
\affiliation{
\country{}
\institution{College of CS, Sichuan University}
}
\email{qjzhao@scu.edu.cn}

%%
%% By default, the full list of authors will be used in the page
%% headers. Often, this list is too long, and will overlap
%% other information printed in the page headers. This command allows
%% the author to define a more concise list
%% of authors' names for this purpose.
\renewcommand{\shortauthors}{Zhang et al.}

%%
%% The abstract is a short summary of the work to be presented in the
%% article.
\begin{abstract}
RGB-D salient object detection (SOD) recently has attracted increasing research interest by benefiting conventional RGB SOD with extra depth information. However, existing RGB-D SOD models often fail to perform well in terms of both efficiency and accuracy, which hinders their potential applications on mobile devices and real-world problems. An underlying challenge is that the model accuracy usually degrades when the model is simplified to have few parameters.  
To tackle this dilemma and also inspired by the fact that depth quality is a key factor influencing the accuracy, we propose a novel depth quality-inspired feature manipulation (DQFM) process, which is efficient itself and can serve as a gating mechanism for filtering depth features to greatly boost the accuracy. DQFM resorts to the alignment of low-level RGB and depth features, as well as holistic attention of the depth stream to explicitly control and enhance cross-modal fusion.   
We embed DQFM to obtain an efficient light-weight model called \ourmodel, where we also design a tailored depth backbone and a two-stage decoder for further efficiency consideration.  
Extensive experimental results demonstrate that our \ourmodel~achieves state-of-the-art accuracy when comparing to existing non-efficient models, and meanwhile runs at 140ms on CPU (2.2$\times$ faster than the prior fastest efficient model) with only $\sim$8.5Mb model size (14.9\% of the prior lightest). Our code will be available at \url{https://github.com/zwbx/DFM-Net}.

%Since the rise in mobile device, the efficient model for RGB-D task is desirable. However, most existing models often fail to perform well in terms of time latency, model size, and accuracy at the same time, for the accuracy always tends to decrease when parameters/time latency is reduced. To break through this dilemma, we proposed a efficient multi-granularity multi-hierarchy tunable gate to explicitly filter the depth features according to its quality, which can greatly boost the accuracy of the results. Thanks to the powerful gate, we just need to embed it into a minimalist pipeline to make accurate predictions, yielding a efficient network termed as \ourmodel. Specifically, the tunable gate can weight proportion where the depth features fused with RGB features according its quality with very low time-space overhead. In addition, we design a light-weight depth backbone and a two-stage decoder as parts of a minimalist pipeline. Extensive experiments demonstrate that the proposed method achieves state-of-the-art performance comparable with non-efficient network. Remarkably, \ourmodel~  also runs at 140ms in CPU(4.6$\times$ faster than prior fastest) for 256$\times$256 RGB-D inputs with only 8.5Mb model size (15\% of the prior lightest). Codes will be made available soon.
\end{abstract}

%%
%% The code below is generated by the tool at http://dl.acm.org/ccs.cfm.
%% Please copy and paste the code instead of the example below.
%%
% \begin{CCSXML}
% <ccs2012>
%  <concept>
%   <concept_id>10010520.10010553.10010562</concept_id>
%   <concept_desc>Computer systems organization~Embedded systems</concept_desc>
%   <concept_significance>500</concept_significance>
%  </concept>
%  <concept>
%   <concept_id>10010520.10010575.10010755</concept_id>
%   <concept_desc>Computer systems organization~Redundancy</concept_desc>
%   <concept_significance>300</concept_significance>
%  </concept>
%  <concept>
%   <concept_id>10010520.10010553.10010554</concept_id>
%   <concept_desc>Computer systems organization~Robotics</concept_desc>
%   <concept_significance>100</concept_significance>
%  </concept>
%  <concept>
%   <concept_id>10003033.10003083.10003095</concept_id>
%   <concept_desc>Networks~Network reliability</concept_desc>
%   <concept_significance>100</concept_significance>
%  </concept>
% </ccs2012>
% \end{CCSXML}

% \ccsdesc[500]{Computer systems organization~Embedded systems}
% \ccsdesc[300]{Computer systems organization~Redundancy}
% \ccsdesc{Computer systems organization~Robotics}
% \ccsdesc[100]{Networks~Network reliability}

%%
%% Keywords. The author(s) should pick words that accurately describe
%% the work being presented. Separate the keywords with commas.

%% A "teaser" image appears between the author and affiliation
%% information and the body of the document, and typically spans the
%% page.

% \begin{teaserfigure}
%   \includegraphics[width=\textwidth]{sampleteaser}
%   \caption{Seattle Mariners at Spring Training, 2010.}
%   \Description{Enjoying the baseball game from the third-base
%   seats. Ichiro Suzuki preparing to bat.}
%   \label{fig:teaser}
% \end{teaserfigure}

%%
%% This command processes the author and affiliation and title
%% information and builds the first part of the formatted document.
\maketitle

\section{Introduction}\label{sec:introduction}

Salient object detection (SOD) aims to locate image regions that attract master human visual attention. It is useful in many down-stream tasks \eg object segmentation \cite{SaliencyAwareVO}, medical segmentation \cite{fan2020pra,ji2021pnsnet}, tracking \cite{2019Non}, image/video compression \cite{2010A}. Owing to the powerful representation ability of deep learning, great progresses have been made in SOD in recent years, but most of them use only RGB images as input to detect salient objects \cite{RGBsurvey,deng2021re,jiang2020light}. This is unavoidable to encounter challenges in complex scenarios, such as cluttered or low-contrast background. 

With the popularity of depth sensors/devices, RGB-D SOD has become a hot research topic \cite{BBSNet,fu2021siamese,UCNet-TPAMI,HDFNet,zhang2021bilateral,zhai2020bifurcated}, because additional useful spatial information embedded in depth maps could serve as a complementary cue for more robust detection~\cite{RGBDsurvey}. Meanwhile, depth data is already widely available on many mobile devices~\cite{D3Net}\eg{}modern smartphones like Huawei Mate 40 Pro, iPhone 12 Pro, Samsung Galaxy S20+. This has opened up a new range of applications for the RGB-D SOD task. Unfortunately, the time-space consumption of existing approaches~\cite{JLDCF,D3Net,S2MA,UCNet,SSF,cmMS,DPA,CPFP,DRMA} is still too high, hindering their further applications on mobile devices and real-world problems. Therefore, an efficient and accurate RGB-D SOD model is highly desirable.

\begin{figure}
  \centering
 \centerline{\epsfig{figure=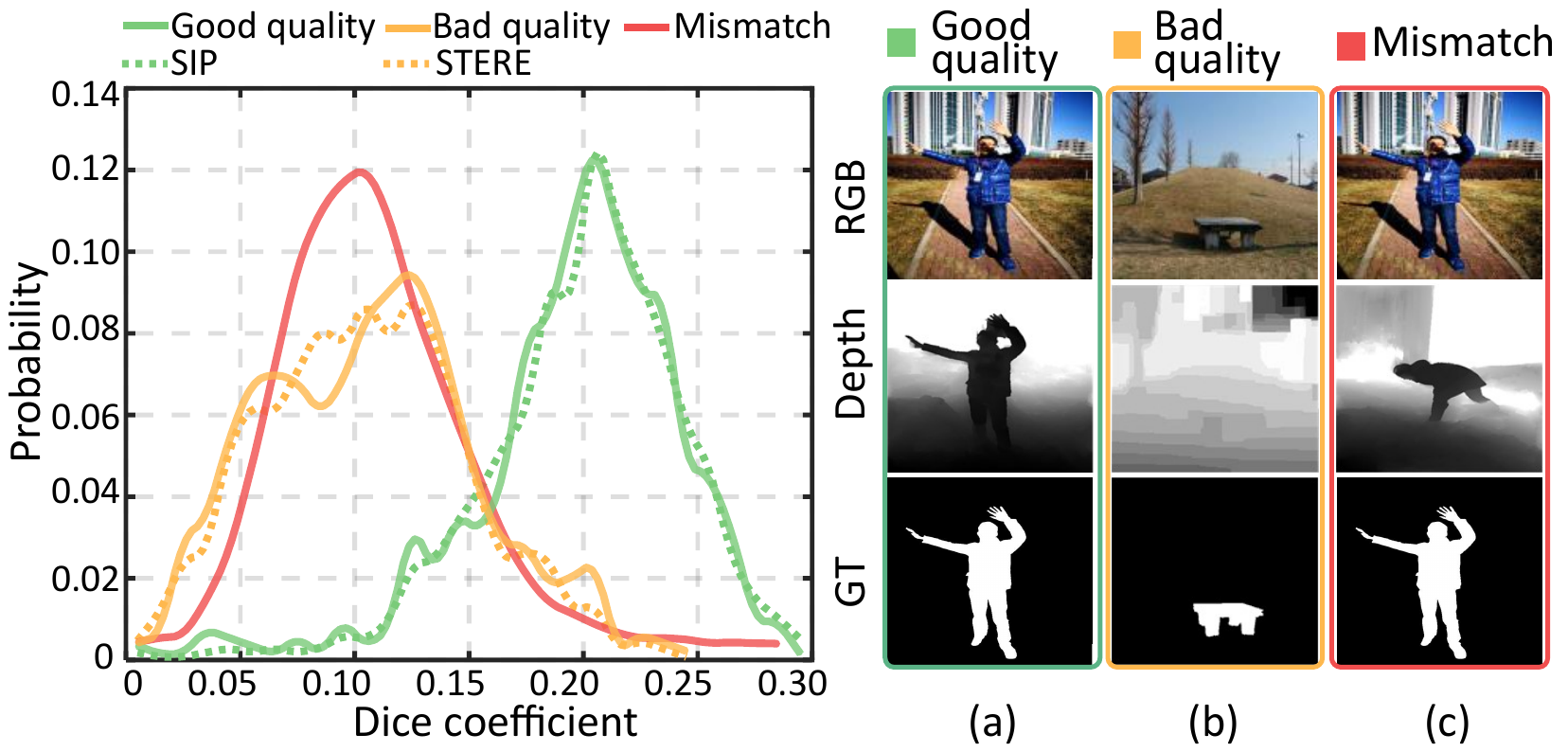,width=0.48\textwidth}}\vspace{-0.3cm}
\caption{Average probability distributions (solid curves) of edge Dice coefficients computed from ``Good quality'', ``Bad quality'', and ``Mismatch'' subsets (10 times random sampling), as well as the probability distributions (dash curves) of the whole STERE and SIP datasets. The right images show three examples from the three subsets, respectively. }\vspace{0cm}
\label{fig:distribution}
\end{figure}

An efficiency-oriented model should aim to significantly decrease the number of operations and memory needed while retaining decent accuracy~\cite{mobilenetV2}. In light of this, several efficiency-oriented RGB-D SOD models have emerged with specific considerations. For example, \cite{A2dele,CoNet} adopt a depth-free inference strategy for fast speed, while depth cues are only utilized in the training phase. \cite{DANet,PGAR,ATSA} choose to design efficient cross-modal fusion modules \cite{DANet} or light depth backbones \cite{PGAR,ATSA}. However, these methods achieve high efficiency at the sacrifice of state-of-the-art accuracy. One underlying challenge is that their accuracy is likely to degrade when the corresponding models are simplified or reduced. % as shown in Fig~\ref{fig:benchmark}.   

\textbf{Motivation.} We notice that unstable quality of depth is one key factor which largely influences the accuracy, as mentioned in~\cite{DPA,D3Net,depthconfidence}. However, very few existing models explicitly take this issue into consideration. We also argue that depth quality is difficult to determine solely according to a depth map \cite{depthconfidence,DPA}, because it is tough to judge whether a salient region in the depth map belongs to noise or a target object, as exemplified in Fig. \ref{fig:distribution} (b). Since RGB-D SOD concerns two paired images as input,\ie{}an RGB image and a depth map, our observation is that a high-quality depth map usually has some boundaries \emph{well-aligned} to the corresponding RGB image. We call this observation ``boundary alignment'' (BA). To validate 
such a BA observation, we randomly choose 50 paired samples (tagged as ``Good quality'' as shown in Fig. \ref{fig:distribution}) from SIP~\cite{D3Net} dataset, and also 50 ``Bad quality'' samples from STERE \cite{STERE} dataset. Choosing such two datasets is based on the general observations of previous works \cite{D3Net,JLDCF}. Additionally, we construct a new set of samples from the ``Good quality'' set, tagged as ``Mismatch'', by randomly mismatching the RGB and depth images of the ``Good quality'' set, to see if this behavior can be reflected by BA. Note that this behavior actually causes no changes to individual RGB or depth images, therefore having no impact to a depth quality measurement (e.g., \cite{depthconfidence}) that is dependant only on depth itself (often called no-reference metric~\cite{IQA}).  To determine boundary alignment, an off-the-shelf edge detector \cite{BDCN} is used to obtain two edge maps from RGB and depth, respectively. We calculate their 
Dice coefficients \cite{VNet} as a measure of BA. The probability distributions (average of 10 times random sampling) of Dice coefficients are shown in Fig. \ref{fig:distribution}, where the three sets of samples correspond to different colors. We can see that BA seems a strong evidence for the depth quality issue, and meanwhile for the ``Mismatch'' set, its Dice coefficients are generally lower than those of the ``Good quality'' set.
%answering a question that ``Is this depth map a high-quality depth map of the RGB image?''. For the ``Mismatch'' set, we can see that the Dice scores are generally much lower than the ``Good quality'' set. 

Inspired by the above fact that the alignment of low-level RGB and depth features can somewhat reflect depth quality, we propose a new depth quality-inspired feature manipulation (DQFM) process. The intuition behind DQFM is to assign \emph{lower weights} to depth features if the quality of depth is \emph{bad}, effectively avoiding injecting noisy or misleading depth features to improve detection accuracy for efficient models. We also augment DQFM with depth holistic attention, in order to \emph{enhance} depth features when the depth quality is judged to be \emph{good}. With the help of DQFM, we explicitly control and enhance the role of depth features during cross-modal fusion. Further in this paper, we embed DQFM into an encoder-decoder framework to obtain an efficient light-weight model called \ourmodel~\emph{(Depth Feature Manipulation Network)}, where a tailored depth backbone and a two-stage decoder are designed for efficiency consideration. The main contributions are summarized as follows:
\begin{itemize} 
    \item We propose an efficient depth quality-inspired feature manipulation (DQFM) process, to explicitly control and enhance depth features during cross-modal fusion. DQFM avoids injecting noisy or misleading depth features, and can effectively improve detection accuracy for efficient models.
    
    \item Benefited from DQFM, we propose an efficient light-weight model \ourmodel~\emph{(Depth Feature Manipulation Network)},  which has a tailored depth backbone and a two-stage decoder.
    
    \item Compared to 15 state-of-the-art (SOTA) models, \ourmodel~is able to achieve superior accuracy, meanwhile running at 7 FPS on CPU (2.2$\times$ faster than the prior fastest model) with only $\sim$8.5Mb model size (14.9\% of the prior smallest). 
\end{itemize}

% \vspace{-17pt}
\section{Related Work}\label{sec:relatedwork}

%-------------------------------------------------------------------------
The utilization of RGB-D data for SOD has been extensively explored for years. Based on the goal of this paper, in this section, we review general RGB-D SOD methods, as well as previous works on efficient models and depth quality analyses.

% \vspace{-15pt}
\subsection{General RGB-D SOD Methods}

% Based on the scope of this paper, we divide existing deep-based models into two types according to how they extract RGB and depth features, namely: parallel independent encoders (Fig. \ref{class} (a)), and tailor-maid sub-networks from depth to RGB (Fig. \ref{class} (b)).
% Recent growing interests in mobile vision applications have generated a high demand for efficient saliency object detection. Several attempts have been made to design a lightweight framework

% A general RGB-D SOD model often runs at device with modern GPU, which mainly focus on how to improve the prediction accuracy and take less consideration about time-space overhead.

Traditional methods mainly rely on hand-crafted features \cite{RGBD135,2013An,2012Context,2015Exploiting}. Lately, deep learning-based methods have made great progress and gradually become a mainstream \cite{PCF,HDFNet,UCNet,DRMA,JLDCF,SSF,cmMS,PDNet,CPFP,BBSNet,MMCI,CoNet,D3Net,DANet}. Qu\etal{}\cite{QuRGBD} first introduced CNNs to infer object saliency from RGB-D data.
Zhu\etal{}\cite{PDNet} designed a master network to process RGB data, together with a sub-network for depth data, and then incorporated depth features into the master network. 
Fu \etal{}\cite{JLDCF} utilized a Siamese network for simultaneous RGB and depth feature extraction, which discovers the commonality between these two views from a model-based perspective. Zhang \etal{}\cite{UCNet} proposed a probabilistic network via conditional variational auto-encoders to model human annotation uncertainty. 
Zhang \etal{}\cite{SSF} proposed a complementary interaction fusion framework to locate salient objects with fine edge details. 
Liu \etal{}\cite{S2MA} introduced a selective self-mutual attention mechanism that can fuse attention learned from both modalities.   
Li \etal{}\cite{CMWNet} designed a cross-modal weighting network to encourage cross-modal and cross-scale information fusion from low-, middle- and high-level features. 
Fan \etal{}\cite{BBSNet} adopted a bifurcated backbone strategy to split multi-level features into student and teacher ones, in order to suppress distractors within low-level layers. 
Pang \etal{}\cite{HDFNet} provided a new perspective to utilize depth information, in which the depth and RGB features are combined to generate region-aware dynamic filters to guide the decoding in the RGB stream. 
Li \etal{}\cite{cmMS} proposed a cross-modality feature modulation module that enhances feature representations by taking depth features as prior. 
Luo \etal{}\cite{CAS-GNN} utilized graph-based techniques to design a network architecture for RGB-D SOD. 
Ji \etal{}\cite{CoNet} proposed a novel collaborative learning framework, where multiple supervision signals are employed, yielding a depth-free inference method. 
Zhao \etal{}\cite{DANet} designed a single stream network to directly take a depth map as the fourth channel of an RGB image, and proposed a depth-enhanced dual attention module.
A relatively complete survey on RGB-D SOD can be found in \cite{RGBDsurvey}.

Despite that encouraging detection accuracy has been obtained by the above RGB-D SOD methods, most of them have heavy models and are computationally expensive. 

\begin{figure*}
  \centering
 \centerline{\epsfig{figure=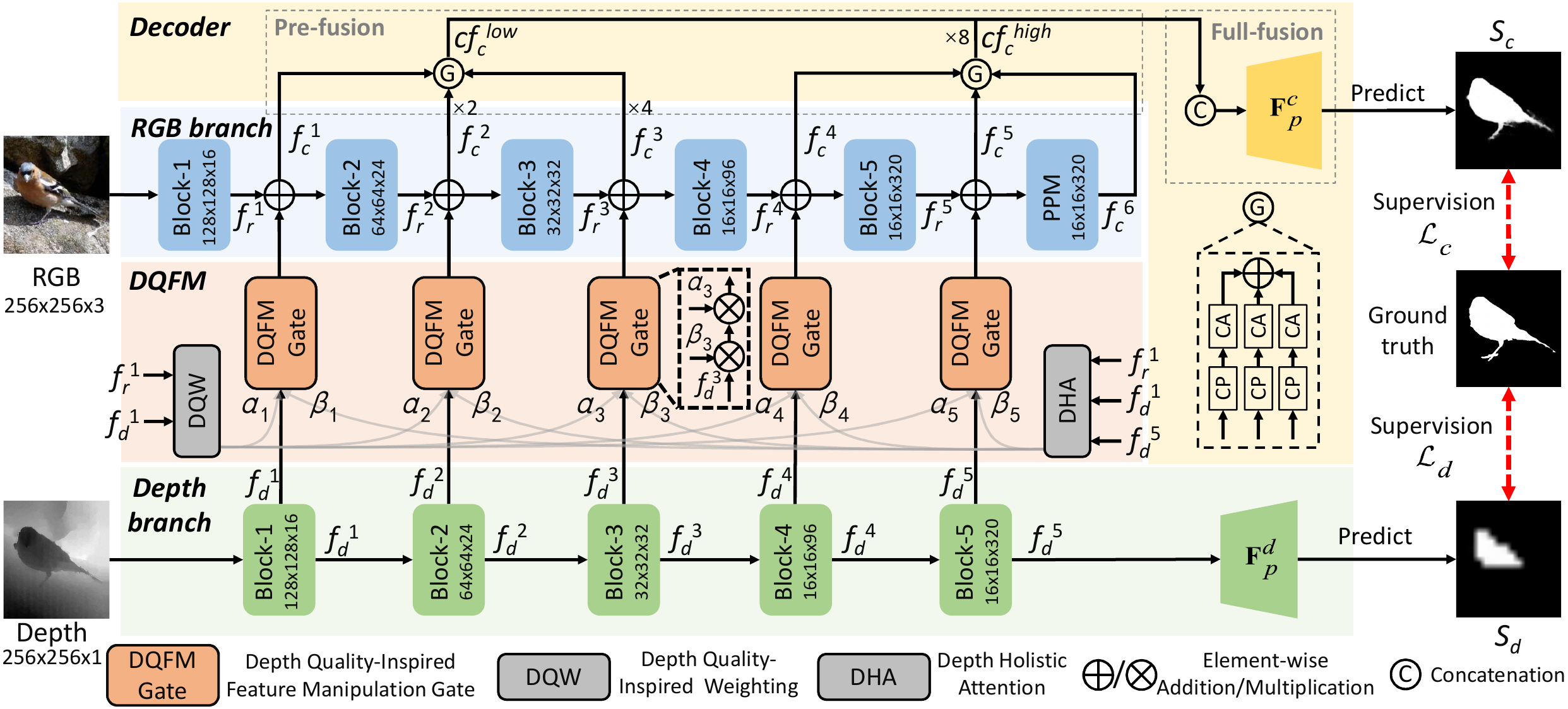,width=1\textwidth}}\vspace{-0.3cm}
\caption{Block diagram of the proposed \ourmodel. Best view in color.}\vspace{-0.2cm}
\label{fig:structure}
\end{figure*}

\subsection{Efficient RGB-D SOD Methods}
Besides the above-mentioned methods, several recent methods attempt to take model efficiency into consideration\footnote{The concept of ``efficient model'' is hard to define. Here and also in the following experiment section, we consider that efficient models should be less than 100Mb.}. 
Specific techniques are used to reduce high computation brought by multi-modal feature extraction and fusion. 
Piao \etal{}\cite{A2dele} employed knowledge distillation for a depth distiller, which aims at transferring depth knowledge obtained from the depth stream to the RGB stream, thus allowing a depth-free inference framework. Chen \etal{}\cite{PGAR} constructed a tailored depth backbone to extract complementary features. Such a backbone is much more compact and efficient than classic backbones\eg{}ResNet \cite{ResNet} and VGGNet \cite{vgg}. Besides, the method adopts a coarse-to-fine prediction strategy that simplifies the top-down refinement process. The utilized refinement module is in a recurrent manner, further reducing model parameters. 

\subsection{Depth Quality Analyses in RGB-D SOD}
% Since the quality of depth often affects model performance, a few researchers have considered the depth quality issue in RGB-D SOD, so as to alleviate the impact of low-quality depth.
% Cong\etal{}~\cite{depthconfidence} first proposed a no-reference depth quality metric \cite{IQA} to alleviate the contamination of low-quality depth. Later, Fan\etal{}~\cite{D3Net} used two networks sharing the same structure to process RGB-D input and depth input individually. 
% Depth quality is then evaluated by comparing between the results from the two networks. 
% More recently, Cong\etal{}~\cite{DPA} proposed to compute quality scores by comparing between a ground truth map and a thresholded depth map. Such quality scores are calculated using IoU and recall, and are then used as labels to train a perceptron for depth quality assessment. Meanwhile, the obtained quality scores guide subsequent multi-modal fusion.
% Different form the above methods, our \ourmodel~exploits the depth quality issue from a new perspective, and leverages RGB as reference to induce depth quality.
Since the quality of depth often affects model performance, a few researchers have considered the depth quality issue in RGB-D SOD, so as to alleviate the impact of low-quality depth. 
As early attempts,  some works proposed to conduct depth quality assessment from a global perspective and obtained a quality score.
Cong\etal{}~\cite{depthconfidence} first proposed a no-reference depth quality metric \cite{IQA} to alleviate the contamination of low-quality depth. Later, Fan\etal{}~\cite{D3Net} used two networks sharing the same structure to process RGB-D input and depth input individually. 
Depth quality is then evaluated by comparing between the results from the two networks. 
Cong\etal{}~\cite{DPA} proposed to predict the depth quality score via a perceptron with high-level RGB and depth features as input. Such a perceptron was trained by scores calculated by comparing thresholded depth maps with ground truth. 

Instead of a single quality score, more recently, spatial quality evaluation of depth maps was also considered, in order to find valuable depth region.
Wang\etal{}~\cite{DQSF} designed three hand-crafted features to excavate depth following multi-scale methodology. Chen\etal{}~\cite{DQSD} proposed to locate the ``Most Valuable Depth Regions'' of depth by
comparing pseudo GT generated from a sub-network with RGB-D as input, with two saliency maps generated from two sub-networks with RGB/depth as input.

Different from the above existing methods that have high time-space complexity, our depth quality assessment is much more efficient and is more suitable to benefit a light-weight model. Besides, our quality module is end-to-end trainable and is also unsupervised (\cite{DPA,DQSD} are supervised for quality estimation).

\section{Methodology}
%In this section, we first provide an overview of the proposed \ourmodel~in Section~\ref{sec:overview}. Then, DQFM, which consists of two parts, namely DQW (XXX) and DHA (XXXX) is described in Section~\ref{sec:DQFM}, and the tailored depth backbone is detailed in Section~\ref{sec:DepthBackbone}. Section~\ref{sec:Decoder} gives details of the designed two-stage decoder. The loss function will be detailed in Section~\ref{sec:loss}.
%\subsection{Overview}\label{sec:overview}

\subsection{Overview}

Fig. \ref{fig:structure} shows the block diagram of the proposed \ourmodel, which consists of encoder and decoder parts. For efficiency consideration, our encoder part follows the design in \cite{BBSNet}, where the RGB branch is simultaneously responsible for RGB feature extraction and cross-modal fusion between RGB and depth features.
The decoder part, on the other hand, conducts simple two-stage fusion to generate the final saliency map. 
More specifically, the encoder consists of an RGB-related branch which is based on MobileNet-v2~\cite{ResNet}, a depth-related branch which is a tailored efficient backbone, and also the proposed DQFM. Both branches lead to five feature hierarchies, and the output stride is 2 for each hierarchy except 1 for the last hierarchy. The extracted depth features at a certain hierarchy, after passing through a DQFM gate, are then fused into the RGB branch by simple element-wise addition before being sent to the next hierarchy.
Besides, in order to capture multi-scale semantic information, we add a PPM (pyramid pooling module \cite{ppm}) at the end of the RGB branch.
Note that in practice, the DQFM gate contains two successive operations, namely depth quality-inspired weighting (DQW) and depth holistic attention (DHA). 

Let the features from the five RGB/depth hierarchies be denoted as $f_{m}^{i}~(m\in\{r,d\},i=1,...,5)$, the fused features be denoted as $f_{c}^{i}~(i=1,...,5)$, and the features from PPM be denoted as $f_{c}^{6}$. The aforementioned cross-modal feature fusion can be expressed as:
\begin{gather}\label{eq:1}
 f_{c}^{i}= f_{r}^{i} + (\alpha_{i}\otimes\beta_{i}\otimes{f}_{d}^{i}),
\end{gather}
where $\alpha_{i}$ and $\beta_{i}$ are computed by DQW and DHA to manipulate  depth features $f_{d}^{i}$ to be fused, and $\otimes$ denotes element-wise multiplication\footnote{If the multipliers' dimensions are different, before element-wise multiplication, the one with less dimension will be replicated to have the same dimension of the other(s).}. After encoding as shown in Fig. \ref{fig:structure}, $f_{c}^{i}~(i=1,...,5)$ and $f_{c}^{6}$ are fed to the subsequent decoder part. 

\begin{figure}
  \centering
 \centerline{\epsfig{figure=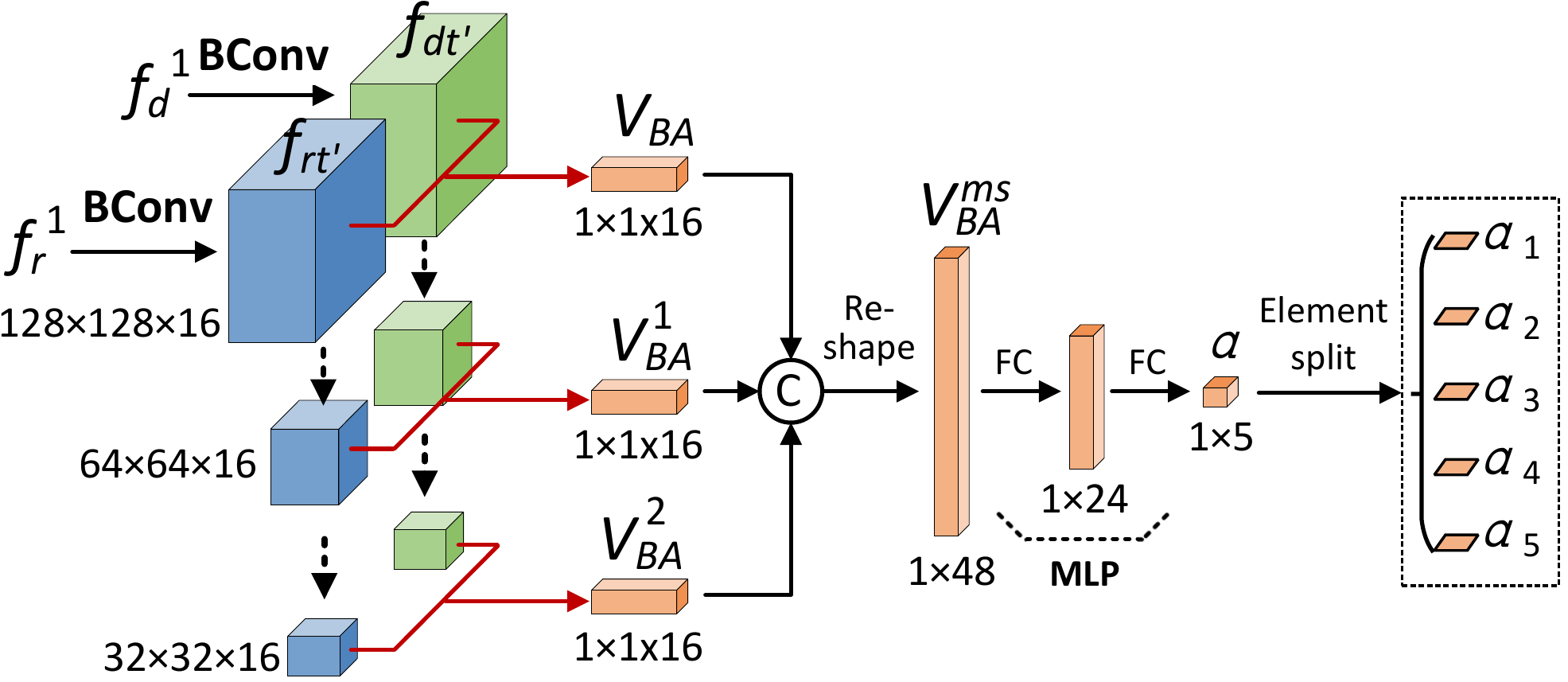,width=0.48\textwidth}}\vspace{-0.2cm}
\caption{Architecture of DQW (depth quality-inspired weighting). The red line arrows indicate the computation of Eq.~\ref{eq:3}. The dash line arrows indicate the max pooling with stride 2.}\vspace{0cm}
\label{fig:dqw}
\end{figure} 

\subsection{Depth Quality-Inspired Feature Manipulation (DQFM)}\label{sec:DQFM}

DQFM consists of two key components, namely DQW (depth quality-inspired weighting) and DHA (depth holistic attention), which generate $\alpha_{i}$ and $\beta_{i}$ in Eq.  \ref{eq:1}, respectively. $\alpha_{i}\in\mathbb{R}^{1}$ is a scalar determining \emph{``how much''} depth features are involved, while $\beta_{i}\in\mathbb{R}^{k \times k}$ ($k$ is the feature size at hierarchy $i$) is a spatial attention map, deciding \emph{``what regions''} to focus in the depth features. Below we describe the internal structures of DQW and DHA.

\textbf{Depth Quality-Inspired Weighting (DQW).} Inspired by the aforementioned BA observation in Sec. \ref{sec:introduction}, as shown in Fig.~\ref{fig:dqw}, DQW learns the weighting term $\alpha_{i}$ adaptively from low-level features $f_{r}^{1}$ and $f_{d}^{1}$, because such low-level features characterize image edges/boundaries \cite{Amulet}. 
To this end, we first apply convolutions to $f_{r}^{1}$/$f_{d}^{1}$ to obtain transferred features $f_{rt'}$/$f_{dt'}$, which are expected to capture more edge-related activation:
\begin{gather}
f_{rt'}=\textbf{BConv}_{1 \times1}(f_{r}^{1}),~f_{dt'}=\textbf{BConv}_{1 \times1}(f_{d}^{1}),
\end{gather}
where $\textbf{BConv}_{1 \times 1}(\cdot{})$ denotes a $1 \times 1$ convolution with BatchNorm and ReLU activation. To evaluate low-level feature alignment and inspired by Dice coefficient \cite{VNet}, given edge activation $f_{rt'}$ and $f_{dt'}$, the alignment feature vector $V_{BA}$ that encodes the alignment between $f_{rt'}$ and $f_{dt'}$ is computed as:
\begin{gather}
\label{eq:3}
V_{BA}=\frac{\textbf{GAP}(f_{rt'} \otimes f_{dt'})}{\textbf{GAP}(f_{rt'} + f_{dt'})},
\end{gather}
where $\textbf{GAP}(\cdot{})$ denotes the global average pooling operation, and $\otimes$ means element-wise multiplication. To make $V_{BA}$ robust against slight edge shifting, we propose to compute $V_{BA}$ at multi-scale and concatenate the results to generate an enhanced vector. As shown in Fig.~\ref{fig:dqw}, such multi-scale calculation is conducted by subsequently down-sampling the initial features $f_{rt'}$/$f_{dt'}$ by max pooling with stride 2, and then computing $V_{BA}^{1},V_{BA}^{2}$ the same as Eq.~\ref{eq:3}. Suppose $V_{BA}$, $V_{BA}^{1}$, and $V_{BA}^{2}$ are the alignment feature vectors computed from three scales as shown in Fig.~\ref{fig:dqw}, the enhanced vector $V_{BA}^{ms}$ is computed as:
\begin{gather}
V_{BA}^{ms}=[V_{BA},V_{BA}^{1},V_{BA}^{2}],
\end{gather}
where $[\cdot]$ denotes channel concatenation. Next, two fully connected layers are applied to derive $\alpha \in \mathbb{R}^{5}$ from $V_{BA}^{ms}$:
\begin{gather}
\alpha = \textbf{MLP}({V_{BA}^{ms}}),
\end{gather}where $\textbf{MLP}(\cdot{})$ denotes a two-layer perceptron with the Sigmoid function at the end. Thus the vector $\alpha$ obtained contains $\alpha_{i} \in (0,1)~(i={1,2...,5})$ as its elements. Notably, here we adopt different weighting terms for different hierarchies rather than an identical one. The effectiveness of this strategy is validated in Sec.~\ref{sec:ablation}.

\begin{figure}
  \centering
 \centerline{\epsfig{figure=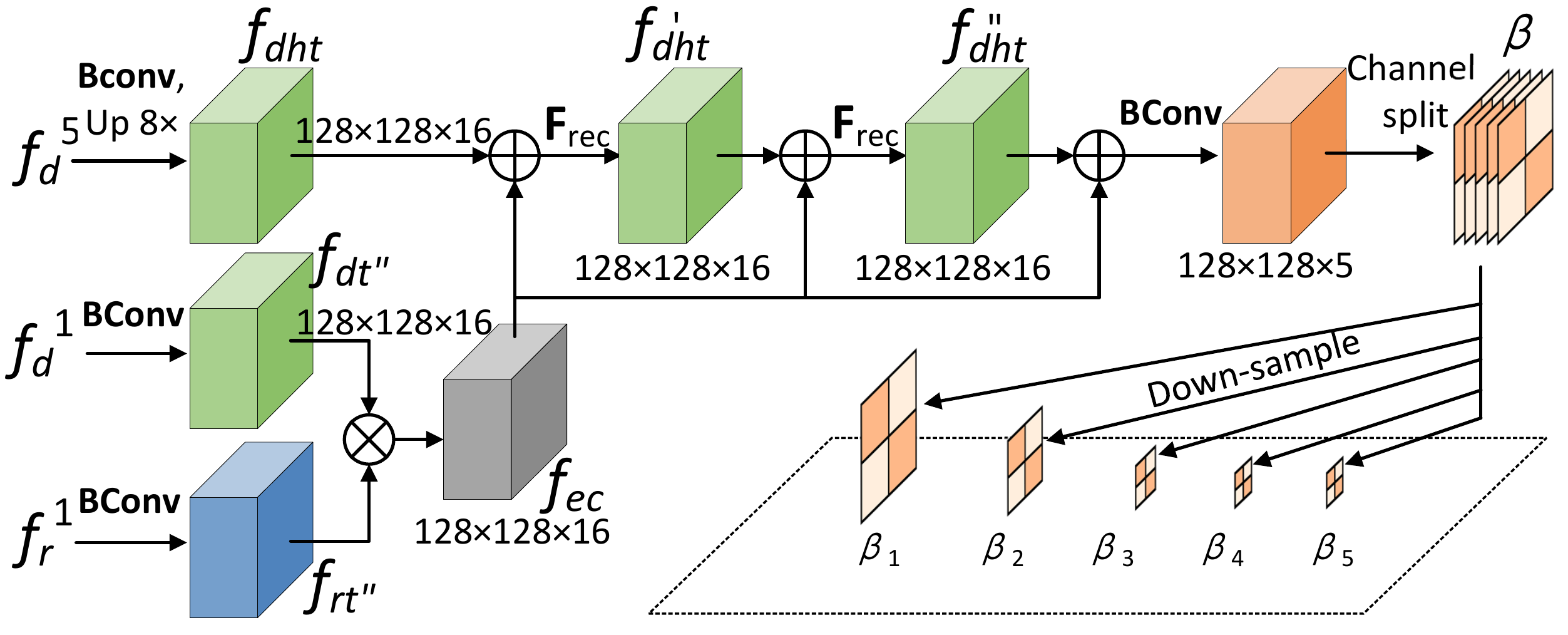,width=0.48\textwidth}}\vspace{0cm}
\caption{Architecture of DHA (depth holistic attention).}\vspace{0cm}
\label{fig:dha}
\end{figure} 
\begin{figure}
  \centering
 \centerline{\epsfig{figure=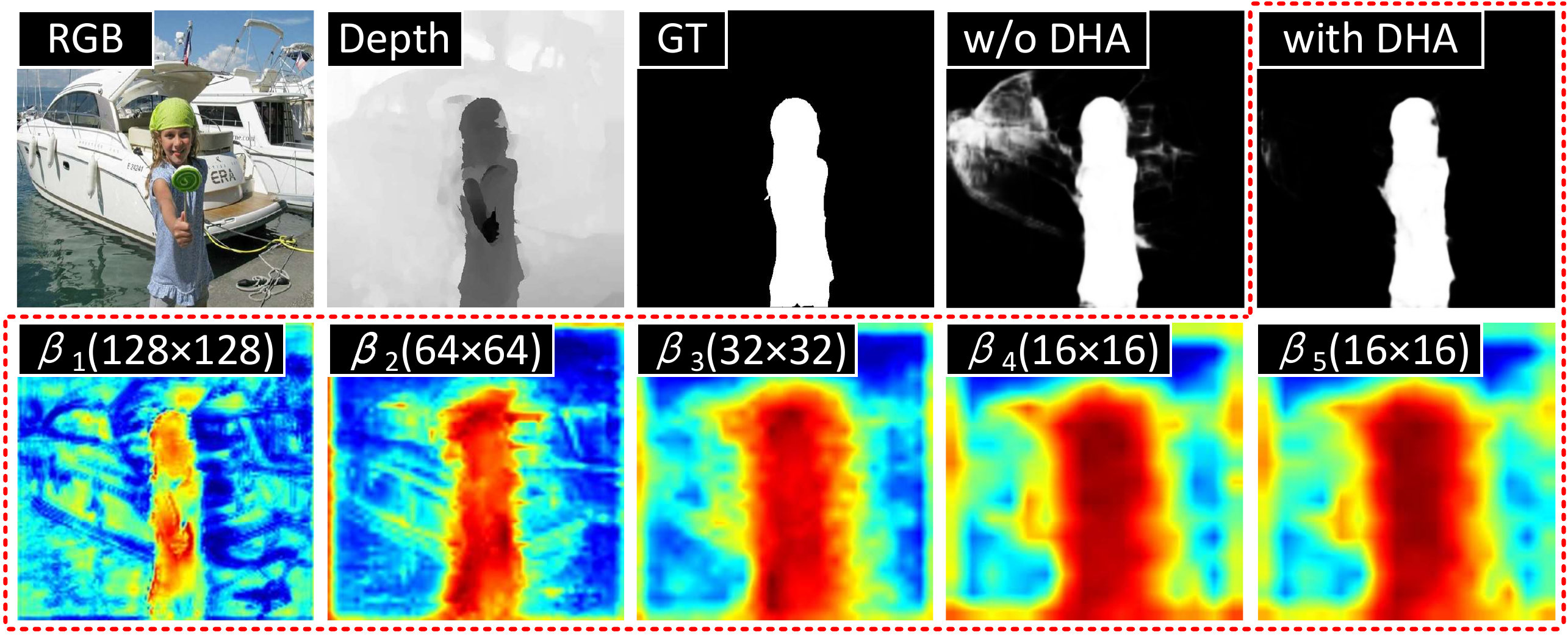,width=0.48\textwidth}}\vspace{-0.3cm}
\caption{Visualization of holistic attention maps $\beta_1 \sim \beta_5$, as well as the resulting saliency maps with and without DHA.}\vspace{0cm}
\label{fig:qualitative_dha}
\end{figure} 
\begin{figure*}[h]
  \centering
 \centerline{\epsfig{figure=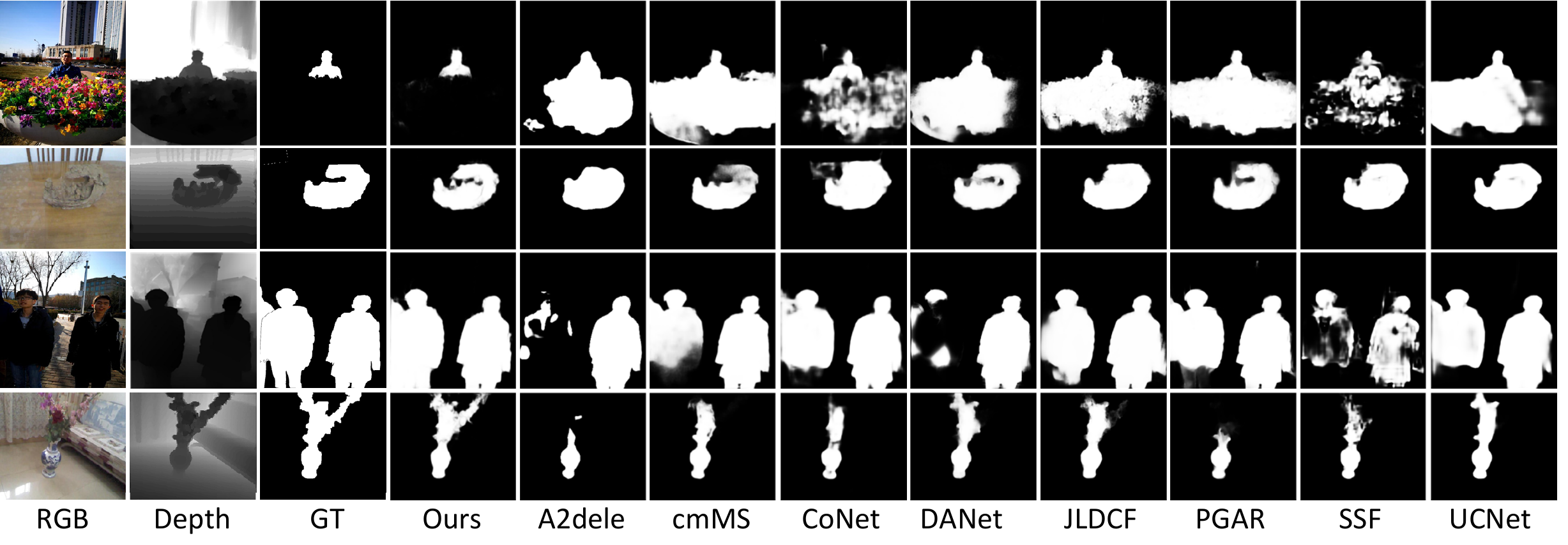,width=1.0\textwidth}}\vspace{-0.3cm}
\caption{Qualitative comparisons of our method (\ourmodel) with SOTA RGB-D SOD models.}\vspace{0cm}
\label{fig:visual_results}
\end{figure*}

\textbf{Depth Holistic Attention (DHA).}\label{sec:DHA} Depth holistic attention (DHA) enhances depth features spatially, by deriving holistic attention map $\beta_i$ from the depth stream. Technically as in Fig. \ref{fig:dha}, we first utilize the highest-level features $f_{d}^{5}$ from the depth stream to locate coarse salient regions (with supervision signals imposed as shown in Fig. ~\ref{fig:structure}). To facilitate subsequent pixel-wise operations, we compress and then up-sample $f_{d}^{5}$ into $f_{dht}$, which has the same dimension as $f_{r}^{1}$/$f_{d}^{1}$, formulated as:
\begin{gather}
f_{dht} = \textbf{F}_{UP}^{8}( \textbf{BConv}_{1 \times1}(f_{d}^{5})),
\end{gather}
where $\textbf{F}_{UP}^{8}$ means 8$\times$ bilinear up-sampling. Then we combine low-level RGB and depth features to recalibrate $f_{dht}$. Similar to the computation of $V_{BA}$, we first transfer $f_{r}^{1}$/$f_{d}^{1}$ to $f_{rt''}$/$f_{dt''}$ as in DQW. The resulting features are element-wisely multiplied to generate features $f_{ec}$, which emphasizes common edge-related activation. To better model long-range dependencies across low-level and high-level features while maintaining efficiency for DHA, we employ the max pooling operation and dilated convolution to rapidly increase receptive fields. The recalibration process is defined as:
\begin{gather}
\textbf{F}_{rec}(f_{dht}) = \textbf{F}_{UP}^{2}\Big(\textbf{DConv}_{3 \times 3}\big(\textbf{F}_{DN}^{2}(f_{dht} + f_{ec})\big)\Big),
\end{gather}
where $\textbf{F}_{rec}(\cdot)$ denotes once recalibration process. $\textbf{DConv}_{3 \times 3}(\cdot{})$ denotes the $3 \times 3$ dilated convolution with stride 1 and dilation rate 2, followed by BatchNorm and ReLU activation. $\textbf{F}_{UP}^{2}(\cdot) /  \textbf{F}_{DN}^{2}(\cdot)$ denotes bilinear up-sampling/down-sampling operation to $2/(\frac{1}{2})$ times the original size. As a trade-off between performance and efficiency, we conduct recalibration twice as below:
\begin{gather}
f'_{dht} =\textbf{F}_{rec}(f_{dht}),~f''_{dht} =\textbf{F}_{rec}(f'_{dht}),
\end{gather}
where $f'_{dht}$ and $f''_{dht}$ represent the features recalibrated once and twice, respectively. Finally, we combine $ f''_{dht}$ and the previous $f_{ec}$ to achieve holistic attention maps:
\begin{gather}
    \beta = \textbf{BConv}_{3 \times 3}(f_{ec} + f''_{dht}).
\end{gather}
Note that here the ReLU activation in $\textbf{BConv}_{3 \times 3}$ is replaced by the Sigmoid one. Finally we have obtained 5 depth holistic attention maps $\{\beta_1,\beta_2,...,\beta_5\}$ by down-sampling from $\beta$, severing as spatial enhancement terms for the depth hierarchies. A visual case of DHA is shown in Fig.~\ref{fig:qualitative_dha}. One can see that $\beta_1$ is likely to highlight regions around edges, while $\beta_5$ focuses more on the dilated entire object. Generally, irrelevant background clutters in depth features can be somewhat suppressed by multiplying attention maps $\beta_1 \sim \beta_5$.

\begin{table}
	\centering
	\caption{Detailed structure of the proposed tailored depth backbone (TDB), which is based on inverted residual bottleneck blocks (IRB) of MobileNet-V2 \cite{mobilenetV2}. About notations, $t$:~expansion factor of IRB, $c$: output channels, $n$: times the block is repeated, and $s$: stride of hierarchy, which is applied to the first block of the repeating blocks.}\vspace{-0.3cm}
	\label{tab:depthbackbone}
	\small
    %\scriptsize
    %\tiny
	%\renewcommand{\arraystretch}{}
	\renewcommand{\tabcolsep}{2mm}
	\begin{tabular}{c|c|c|c|c|c|c}
		\hline\toprule
 Input& Output & Block &$t$ &$c$ &$n$ &$s$

\\
		\midrule
$256\times256\times1$ &$128\times128\times16$ &IRB &3 &16 &1 &2

 \\
	$128\times128\times16$ &$64\times64\times24$ &IRB &3 &24 &3 &2 

  \\
		$64\times64\times24$ &$32\times32\times32$ &IRB &3 &32 &7 &2
\\
$32\times32\times32$ &$16\times16\times96$ &IRB &2 &96 &3 &2\\
$16\times16\times96$ &$16\times16\times320$ &IRB &2 &320 &1 &1\\
		\bottomrule
		\hline
	\end{tabular}
	\vspace{-8pt}
\end{table}

\subsection{Tailored  Depth Backbone~(TDB)}\label{sec:DepthBackbone}
Usually, depth is less informative than RGB. Hence we consider using a tailored depth backbone (TDB), which is lighter, as a trade-off between efficiency and accuracy. Specifically, we base our TDB on the inverted residual bottleneck blocks (IRB) from MobileNet-V2~\cite{mobilenetV2}, and construct a new smaller backbone with reduced channel numbers and stacked blocks, whose structure is detailed in Tab.~\ref{tab:depthbackbone}. As a result, our TDB is much lighter than previous light-weight backbones\eg{} (Ours: only 0.9Mb, ATSA's \cite{ATSA}: 6.7Mb, PGAR's~\cite{PGAR}: 6.2Mb, MobileNet-V2: 6.9Mb), and meanwhile, its performance is slightly better than MobileNet-V2 (see Sec.~\ref{sec:ablation}). During training, we embed TDB into \ourmodel~without pre-training, and supervision signals are imposed at the end of the backbone to enforce saliency feature learning from depth, as shown in Fig. \ref{fig:structure}. The coarse prediction result obtained from TDB is formulated as:
% But there exist one problem, the representation learning ability of the shallow network decrease due to the reduced parameters. Fortunately, the proposed gate can help light-weight backbone  overcome this limitation to retain prediction accuracy, allowing us to adopt a  lighter backbone. 
%Following~\cite{FastSCNN}, we directly embed it into \ourmodel~  without pre-train and we impose a deep supervision at the tail to enhance its feature representative learning, formulated as:
\begin{gather}
    S_d = \textbf{F}_p^{d}(f_{d}^{5}),
\end{gather}
where $S_d$ means the coarse prediction from TDB, which is supervised by ground truth (GT). $\textbf{F}_p^{d}(\cdot)$ denotes a prediction head consisting of a $1 \times 1$ convolution followed by a BatchNorm layer and Sigmoid activation, and also $16 \times$ bilinear up-sampling to recover the original input size. The effectiveness of the proposed TDB will be validated in Sec.~\ref{sec:ablation} .
% , we design a extreme efficient depth backbone which has the same 5 hierarchy as the RGB backbone, with two-thirds narrower channel and fewer convolution block. Thanks to our proposed gate strategy, our depth backbone  in synergy with our proposed gate to generate a robust depth feature extractor. 

\begin{table*}[htbp]
	\centering
	\caption{Quantitative benchmark results. $\uparrow$/$\downarrow$ for a metric denotes that a larger/smaller value is better. Our results are highlighted in bold. The scores/numbers better than ours are \underline{underlined} (efficient and non-efficient models are labeled separately).}\vspace{0cm}
	%The best scores/numbers are highlighted in \textbf{bold}.}
	\label{tab:benchmark}
	\footnotesize
    %\scriptsize
    %\tiny
	\renewcommand{\arraystretch}{0.12}
	\renewcommand{\tabcolsep}{0.87mm}
	\begin{tabular}{lr|ccccccccccccc|c||cc|c}
		\hline\toprule
		&\multirow{3}{*}{Metric}\centering    &PCF & MMCI & CPFP & DMRA &D3Net   &JL-DCF &UCNet  &SSF  &S2MA &CoNet  &cmMS &DANet &ATSA  &\textbf{\ourmodel$^{*}$} &A2dele &PGAR &\textbf{\ourmodel}  \\&    &\scriptsize CVPR18 &\scriptsize PR19 &\scriptsize CVPR19 &\scriptsize ICCV19 &\scriptsize TNNLS20   &\scriptsize CVPR20 &\scriptsize CVPR20   &\scriptsize CVPR20
		&\scriptsize CVPR20
		&\scriptsize ECCV20   &\scriptsize ECCV20
		&\scriptsize ECCV20 &\scriptsize ECCV20 & \scriptsize \textbf{Ours} &\scriptsize CVPR20&\scriptsize ECCV20 &\scriptsize \textbf{Ours}\\
		&     &\cite{PCF}&\cite{MMCI}&\cite{CPFP} &\cite{DRMA} &\cite{D3Net}   &\cite{JLDCF} &\cite{UCNet}  &\cite{SSF} &\cite{S2MA} &\cite{CoNet}	&\cite{cmMS} &\cite{DANet}  &\cite{ATSA}  &-&\cite{A2dele} &\cite{PGAR}&\textbf{-} \\
		\specialrule{0em}{1pt}{0pt}
		\hline\hline
		\specialrule{0em}{0pt}{1pt}

		%\multirow{4}{*}{\begin{sideways}\cite{ju2014depth}\end{sideways}}
\multirow{3}{*}{\begin{sideways}\textit{}\end{sideways}}

& Size~(Mb)$\downarrow$  &534	&930	&278	&228	&530	&520	&119		&125	&330	&167		&430	&102	&123  &\textbf{93} &57 &62	&\textbf{8.5}
\\

		& CPU~(ms)$\downarrow$  &35762	&46886	&16946	&1521	&851	&7588	&1011		&650	&1900	&667		&946	&1147	&1344  &\textbf{357}    &313    &755 &\textbf{140}
 \\
		& GPU~(FPS)$\uparrow$  &5	&8	&6	&13	&32	&9	&42	&21	&25	&36		&15	&32	&29  &\textbf{70} &\underline{120}	&61 &\textbf{64}
 \\
	    \midrule
		\multirow{4}{*}{\begin{sideways}\textit{SIP}\end{sideways}}
		%\multirow{4}{*}{\begin{sideways}\cite{ju2014depth}\end{sideways}}
		& $S_{\alpha}\uparrow$ &0.842	&0.833	&0.850	&0.806	&0.860	&0.879 &0.875		&0.874	&0.878	&0.858		&0.867	&0.878	&0.864  &\textbf{0.885} &0.829	&0.875 &\textbf{0.883}
\\
		& $F_{\beta}^{\rm max}\uparrow$  &0.838	&0.818	&0.851	&0.821	&0.861	&0.885 &0.879		&0.880	&0.884	&0.867		&0.871	&0.884	&0.873  &\textbf{0.890} &0.834	&0.877&\textbf{0.887}
 \\
		& $E_{\xi}^{\rm max}\uparrow$     &0.901	&0.897	&0.903	&0.875	&0.909	&0.923 &0.919		&0.921	&0.920	&0.913		&0.091	&0.920	&0.911   &\textbf{0.926} &0.889	&0.914&\textbf{0.926}
  \\
		& $\mathcal{M}\downarrow$  &0.071	&0.086	&0.064	&0.085	&0.063	&0.051	&0.051	&0.053	&0.054	&0.063	&0.061	&0.054	&0.058  &\textbf{0.049} &0.070 &0.059&\textbf{0.051}
 \\
		\midrule
		\multirow{4}{*}{\begin{sideways}\textit{NLPR}\end{sideways}}
		%\multirow{4}{*}{\begin{sideways}\cite{peng2014rgbd}\end{sideways}}
		& $S_{\alpha}\uparrow$ &0.874	&0.856	&0.888	&0.899	&0.912	&0.925	&0.920		&0.914	&0.915	&0.908		&0.915	&0.915	&0.907  &\textbf{0.926} &0.890 &0.918	&\textbf{0.923}
 \\
		& $F_{\beta}^{\rm max}\uparrow$    &0.841	&0.815	&0.867	&0.879	&0.897	&\underline{0.916} &0.903	&0.896	&0.902	&0.887		&0.896	&0.903	&0.876  &\textbf{0.912} &0.875	&0.898 &\textbf{0.908}
 \\
		& $E_{\xi}^{\rm max}\uparrow$   &0.925	&0.913	&0.932	&0.947	&0.953	&\underline{0.962} &0.956		&0.953	&0.950	&0.945		&0.949	&0.953	&0.945  &\textbf{0.961} &0.937 &0.948	&\textbf{0.957}
 \\
		& $\mathcal{M}\downarrow$ &0.044	&0.059	&0.036	&0.031	&0.025	&\underline{0.022}	&0.025	&0.026	&0.030  &0.031		&0.027	&0.029	&0.028  &\textbf{0.024}	&0.031&0.028&\textbf{0.026}
 \\
		\midrule
		\multirow{4}{*}{\begin{sideways}\textit{NJU2K}\end{sideways}}
		%\multirow{4}{*}{\begin{sideways}\cite{niu2012leveraging}\end{sideways}}
		& $S_{\alpha}\uparrow$&0.877	&0.858	&0.879	&0.886	&0.900	&0.903	&0.897		&0.899	&0.894	&0.895		&0.900	&0.891	&0.901  &\textbf{0.912}	&0.868&0.906 &\textbf{0.906}
 \\
		& $F_{\beta}^{\rm max}\uparrow$  &0.872	&0.852	&0.877	&0.886	&0.900	&0.903	&0.895		&0.896	&0.889	&0.892	 &0.897 &0.880	&0.893   &\textbf{0.913} &0.872 &0.905	&\textbf{0.910}
 \\
		& $E_{\xi}^{\rm max}\uparrow$&0.924	&0.915	&0.926	&0.927	&0.950	&0.944	&0.936		&0.935	&0.930	&0.937		&0.936	&0.932	&0.921  &\textbf{0.950} &0.914&0.940	&\textbf{0.947}
 \\
		& $\mathcal{M}\downarrow$    &0.059	&0.079	&0.053	&0.051	&0.041	&0.043	&0.043		&0.043	&0.053	&0.047		&0.044	&0.048 	&0.040 &\textbf{0.039} &0.052&0.045&\textbf{0.042}
 \\
		\midrule
		\multirow{4}{*}{\begin{sideways}\textit{RGBD135}\end{sideways}}
		%\multirow{4}{*}{\begin{sideways}\cite{cheng2014depth}\end{sideways}}
		& $S_{\alpha}\uparrow$  &0.842	&0.848	&0.872	&0.900	&0.898	&0.929	&0.934		&0.905	&\underline{0.941}	&0.910		&0.932	&0.904	&0.907	 &\textbf{0.938} &0.884&0.894&\textbf{0.931}
  \\
		& $F_{\beta}^{\rm max}\uparrow$  &0.804	&0.822	&0.846	&0.888	&0.885	&0.919	&0.930		&0.883	&\underline{0.935}	&0.896		&0.922	&0.894	&0.885 &\textbf{0.932} &0.873 &0.879	&\textbf{0.922}
  \\
		& $E_{\xi}^{\rm max}\uparrow$  &0.893	&0.928	&0.923	&0.943	&0.946	&0.968	&\underline{0.976}		&0.941	&{0.973}	&0.945		&0.970	&0.957	&0.952  &\textbf{0.973}	&0.920&0.929 &\textbf{0.972}
\\
		& $\mathcal{M}\downarrow$  &0.049	&0.065	&0.038	&0.030	&0.031	&0.022	& 0.019		&0.025	&0.021	&0.029		&0.020	&0.029	&0.024  &\textbf{0.019} &0.030&0.032	&\textbf{0.021}
 \\
		\midrule
		\multirow{4}{*}{\begin{sideways}\textit{LFSD}\end{sideways}}
% 		\multirow{4}{*}{\begin{sideways}\cite{LFSD~\cite{LFSD}}\end{sideways}}
		& $S_{\alpha}\uparrow$ &0.786	&0.787	&0.828	&0.839	&0.825	&0.862	&0.864	&0.859	&0.837	&0.862	&0.849	&0.845	&0.865
		 &\textbf{0.870} &0.834 & 0.833 &\textbf{0.865}
\\
		& $F_{\beta}^{\rm max}\uparrow$   &0.775	&0.771	&0.826	&0.852	&0.810	&0.866	&0.864		&\underline{0.867} &0.835	&0.859		&\underline{0.869}	&0.846	&0.862  &\textbf{0.866} &0.832&0.831&\textbf{0.864}
\\
		& $E_{\xi}^{\rm max}\uparrow$   &0.827	&0.839	&0.863	&0.893	&0.862	&0.901	&\underline{0.905}		&0.900 &0.873	&\underline{0.907}		&0.896	&0.886	&\underline{0.905} 
		&\textbf{0.903} &0.874 &0.893&\textbf{0.903}
\\
		& $\mathcal{M}\downarrow$  &0.119	&0.132	&0.088	&0.083	&0.095	&0.071	&\underline{0.066}	&\underline{0.066}	&0.094	&0.071		&0.074	&0.083	&\underline{0.064}  &\textbf{0.068}	&0.077&0.093&\textbf{0.072}
\\
		\midrule
		\multirow{4}{*}{\begin{sideways}\textit{STERE}\end{sideways}}
		%\multirow{4}{*}{\begin{sideways}\cite{fan2020rethinking}\end{sideways}}
		& $S_{\alpha}\uparrow$   &0.875	&0.873	&0.879	&0.835	&0.899	&0.905	&0.903		&0.893	&0.890	&0.908		&0.895	&0.892	&0.897	 &\textbf{0.908}&0.885 &\underline{0.903}&\textbf{0.898}
 \\
		& $F_{\beta}^{\rm max}\uparrow$ &0.860	&0.863	&0.874	&0.847	&0.891	&0.901	&0.899		&0.890	&0.882	&0.904	&0.891	&0.881	&0.884  &\textbf{0.904}	&0.885&0.893&\textbf{0.893}
 \\
		& $E_{\xi}^{\rm max}\uparrow$  &0.925	&0.927	&0.925	&0.911	&0.938	&0.946	&0.944		&0.936	&0.932	&0.948		&0.937	&0.930	&0.921  &\textbf{0.948}	&0.935&0.936&\textbf{0.941}
\\
		& $\mathcal{M}\downarrow$   &0.064	&0.068	&0.051	&0.066	&0.046	&0.042	&\underline{0.039}		&0.044	&0.051	&0.040		&0.042	&0.048	&\underline{0.039} 	&\textbf{0.040} &0.043 &\underline{0.044}&\textbf{0.045}
\\
		\bottomrule
		\hline
	\end{tabular}
	\vspace{-8pt}
\end{table*}

\subsection{Two-Stage Decoder}\label{sec:Decoder}
Unlike the popular U-Net \cite{unet} which adopts the hierarchy-by-hierarchy top-down decoding strategy, we propose a simplified two-stage decoder, including pre-fusion and full-fusion, to further improve efficiency. The pre-fusion aims to reduce  feature channels and hierarchies, by channel compression and hierarchy grouping, denoted as ``CP'' and ``G'' in Fig. \ref{fig:structure}. Based on the outputs of pre-fusion, the full-fusion further aggregates low-level and high-level hierarchies to generate the final saliency map.

\textbf{Pre-fusion Stage.} We first use $3\times3$ depth-wise separable convolution\footnote{Instead of the common \textbf{BConv}, depth-wise separable convolution \textbf{DSConv} is used here for large numbers of input channels.} \cite{mobilenet} with BatchNorm and ReLU activation, denoted as $\textbf{DSConv}_{3 \times 3}$, to compress the encoder features ($f_{c}^{i},i=1,2...6$) to a unified channel 16, denoted as ``CP'' in Fig. \ref{fig:structure}. Then we use the well-known channel attention operator \cite{senet} $\textbf{F}_{CA}$ to enhance features by weighting different channels, denoted as ``CA'' in Fig.~\ref{fig:structure}. The above process can be described as:
\begin{gather}
    cf_{c}^{i}=\textbf{F}_{CA}(\textbf{DSConv}_{3 \times 3}(f_{c}^{i})),
\end{gather}
where $cf_{c}^{i}$ denotes the compressed and enhanced features.
To reduce feature hierarchies, inspired by \cite{BBSNet}, we group 6 hierarchies into two (\emph{i.e.}, low-level hierarchy and high-level hierarchy) as below:
\begin{gather}
    cf_{c}^{low}=\sum_{i=1}^{3}\textbf{F}_{UP}^{2^{i-1}}(cf_{c}^{i}),~cf_{c}^{high}=\sum_{i=4}^{6}cf_{c}^{i},
\end{gather}
 where $\textbf{F}_{UP}^{i}$ is bilinear up-sampling to $i$ times the original size.
 
\textbf{Full-fusion Stage.} Since in the pre-fusion stage, the channel numbers and hierarchies are already reduced, in the full-fusion stage, we directly concatenate high-level and low-level hierarchies, and then feed the concatenation to a prediction head to achieve the final  full-resolution prediction map, denoted as:
\begin{gather}
    S_c = \textbf{F}_p^{c}\bigg(\big[cf_{c}^{low},\textbf{F}_{UP}^{8}(cf_{c}^{high})\big]\bigg),
\end{gather}
where $S_c$ is the final saliency map, and $\textbf{F}_{p}^{c}(\cdot)$ indicates a prediction head consisting of two $3\times 3$ depth-wise separable convolutions (followed by BatchNorm layers and ReLU activation), a $3\times 3$ convolution with Sigmoid activation, as well as a $2\times$ bilinear up-sampling to recover the original input size. 
\subsection{Loss Function}\label{sec:loss}
The overall loss $\mathcal{L}$ is composed of the final loss $\mathcal{L}_{c}$ and deep supervision for the depth branch loss $\mathcal{L}_{d}$, formulated as:
\begin{gather}
\mathcal{L}=\mathcal{L}_{c}(S_c,G)+\mathcal{L}_{d}(S_d,G),
\end{gather}
where $G$ denotes the ground truth (GT). Similar to previous works \cite{JLDCF,BBSNet,HDFNet,D3Net}, we use the standard cross-entropy loss for $\mathcal{L}_{c}$ and $\mathcal{L}_{d}$.

\section{Experiments and Results}
\subsection{Datasets and Metrics}
We conduct experiments on six widely used datasets, including NJU2K \cite{NJU2K} (1,985 samples), NLPR \cite{NLPR} (1,000 samples), STERE \cite{STERE} (1000 samples), RGBD135 \cite{RGBD135} (135 samples), LFSD \cite{LFSD} (100 samples), and SIP \cite{D3Net} (929 samples). 
Following previous works \cite{JLDCF,UCNet,CPFP}, we use the same 1,500 samples from NJU2K and 700 samples from NLPR for training, and test on the remaining samples. Four commonly used metrics are used for evaluation, including S-measure ($S_{\alpha}$) \cite{smeasure},  maximum F-measure ($F_{\beta}^{\rm max}$) \cite{fmeasure}, maximum E-measure ($E_{\xi}^{\rm max}$) \cite{emeasure,Fan2018Enhanced}, and mean absolute error (MAE, $\mathcal{M}$) \cite{mae}. For efficiency analysis, we report model size (Mb, Mega-bytes), inference time (ms, millisecond) on CPU, and FPS (frame-per-second) on GPU.

\begin{table*}[htbp]
	\centering
	\caption{Ablation analyses for DQFM, where the effectiveness of DQW and DHA is validated.  Details are in Sec.~\ref{sec:ablation}.}\vspace{-0.3cm}
	\label{tab:ablation1}
	
	\footnotesize
    %\scriptsize
    %\tiny
	\renewcommand{\arraystretch}{0.7}
	\renewcommand{\tabcolsep}{0.44mm}
	\begin{tabular}{c|c|c|c|cccc|cccc|cccc|cccc|cccc|cccc}
		\hline\toprule
		\multirow{2}{*}{\#} 
		&\multirow{2}{*}{DQW}
		&\multirow{2}{*}{DHA}
% 		&\multirow{2}{*}{TB}
		&Size
		&\multicolumn{4}{c|}{\textbf{SIP~\cite{D3Net}}}  &\multicolumn{4}{c|}{\textbf{NLPR~\cite{NLPR}}}
		&\multicolumn{4}{c|}{\textbf{NJU2K~\cite{NJU2K}}}
		&\multicolumn{4}{c|}{\textbf{RGBD135~\cite{RGBD135}}}
		&\multicolumn{4}{c|}{\textbf{LFSD~\cite{LFSD}}}
		&\multicolumn{4}{c}{\textbf{STERE~\cite{STERE}}}\\ &&&(Mb)
% 		&$S_{\alpha}$
% 		& $F_{\beta}^{\rm max}$
% 		&$E_{\xi}^{\rm max}$
% 		&$\mathcal{M}$ 
        &$S_{\alpha}$
	    &$F_{\beta}^{\rm max}$
	    & $E_{\xi}^{\rm max}$ 
		&$\mathcal{M}$  
		&$S_{\alpha}$
	    &$F_{\beta}^{\rm max}$
	    & $E_{\xi}^{\rm max}$ 
		&$\mathcal{M}$  
		&$S_{\alpha}$
	    &$F_{\beta}^{\rm max}$
	    & $E_{\xi}^{\rm max}$ 
		&$\mathcal{M}$  
		&$S_{\alpha}$
	    &$F_{\beta}^{\rm max}$
	    & $E_{\xi}^{\rm max}$ 
		&$\mathcal{M}$  
		&$S_{\alpha}$
	    &$F_{\beta}^{\rm max}$
	    & $E_{\xi}^{\rm max}$ 
		&$\mathcal{M}$  
		&$S_{\alpha}$
	    &$F_{\beta}^{\rm max}$
	    & $E_{\xi}^{\rm max}$ 
		&$\mathcal{M}$

\\
		\midrule
% 		1	&	&	&	&	&0.874	&0.878	&0.919	&0.057	&0.903	&0.879	&0.944	&0.033	&0.891	&0.890	&0.934	&0.05	&0.938	&0.929	&0.975	&0.021	&0.843	&0.843	&0.885	&0.083	&0.873	&0.867	&0.928	&0.057	\\ 2	&\checkmark	&	&	&	&0.875	&0.882	&0.922	&0.055	&0.917	&0.901	&0.954	&0.028	&0.902	&0.903	&0.944	&0.044	&0.938	&0.928	&0.974	&0.020	&0.854	&0.854	&0.893	&0.076	&0.883	&0.878	&0.936	&0.051	\\ 3	&	&\checkmark 	&	&	&0.880	&0.886	&0.923	&0.053	&0.921	&0.904	&0.957	&0.027	&0.904	&0.904	&0.942	&0.042	&0.941	&0.934	&0.978	&0.019	&0.868	&0.863	&0.903	&0.069	&0.890	&0.886	&0.939	&0.048	\\ 
1	&	&&8.409	&0.849	&0.842	&0.897	&0.070	&0.907	&0.884	&0.946	&0.032	&0.890	&0.887	&0.931	&0.052	&0.929	&0.915	&0.968	&0.025	&0.847	&0.841	&0.883	&0.084	&0.880	&0.872	&0.931	&0.054	\\

2	&\checkmark 	& &	8.416	&0.878	&0.884	&0.922	&0.054	&0.918	&0.902	&0.957	&0.027	&0.898	&0.898	&0.938	&0.047	&\textbf{{0.932}}	&0.922	&0.971	&0.022	&0.857	&0.856	&0.896	&0.078	&0.894	&0.886	&0.939	&0.047	\\ 

3	&	&\checkmark 	&8.451		&0.861	&0.861	&0.908	&0.063	&0.918	&0.901	&0.953	&0.028	&0.899	&0.896	&0.940	&0.046	&0.921	&0.911	&0.966	&0.024	&0.857	&0.855	&0.895	&0.076	&0.897	&0.891	&\textbf{{0.942}}	&0.045	\\

\rowcolor{mygray}

4	&\checkmark 	&\checkmark 		&8.459	&\textbf{{0.883}}	&\textbf{{0.887}}	&\textbf{{0.926}}	&\textbf{{0.051}}	&\textbf{{0.923}}	&\textbf{{0.908}}	&\textbf{{0.957}}	&\textbf{{0.026}}	&\textbf{{0.906}}  &\textbf{{0.910}}	&\textbf{{0.947}}	&\textbf{{0.042}}	&0.931	&\textbf{{0.922}}	&\textbf{{0.972}}	&\textbf{{0.021}}	&\textbf{{0.865}}	&\textbf{{0.864}}	&\textbf{{0.903}}	&\textbf{{0.072}}	&\textbf{{0.898}}	&\textbf{{0.893}}	&0.941	&\textbf{{0.045}}	\\

% 		1 & & & & &0.874	&0.878	&0.057		&0.903	&0.879	&0.033	&0.891	&0.89	&0.050	&0.938	&0.929	&0.021	&0.843	&0.843	&0.083		&0.873	&0.867	&0.057

%  \\
%  2 &\checkmark & & & &0.881	&0.887	&0.052		&0.917	&0.901	&0.028		&0.902	&0.905	&0.044		&0.937	&0.929	&0.021		&0.858	&0.860	&0.074		&0.888	&0.885	&0.048

%  \\
         
% 		3 & &\checkmark  & & &0.874	&0.882	&0.054	&0.919	&0.904	&0.027	&0.903	&0.904	&0.043	&0.932	&0.923	&0.021	&0.863	&0.863	&0.070	&0.896	&0.890	&0.045

%   \\
%   4 &\checkmark &\checkmark  & & &\textbf{0.883}	&\textbf{0.887}	&\textbf{0.050}	&\textbf{0.922}	&\textbf{0.908}	&\textbf{0.026}	&\textbf{0.906}	&\textbf{0.908}	&\textbf{0.042}	&0.931	&0.920	&0.022	&\textbf{0.865}	&\textbf{0.867}	&\textbf{0.070}	&0.893	&0.887	&0.046

%         \\\midrule
% 		 5 & &  &\checkmark & &0.849	&0.842	&0.070	&0.907	&0.884	&0.032	&0.890	&0.887	&0.052	&0.929	&0.915	&0.025	&0.847	&0.841	&0.084	&0.880	&0.872	&0.054

%   \\
%   6 &\checkmark & &\checkmark & &0.874	&0.877	&0.057	&0.919	&0.905	&0.028	&0.900	&0.901	&0.046	&0.928	&0.918	&0.024	&0.854	&0.851	&0.078	&0.892	&0.885	&0.048

%  \\ 
% 		7 & &\checkmark  &\checkmark & &0.878	&0.883	&0.054	&0.921	&0.905	&0.027	&0.901	&0.898	&0.044	&0.934	&0.926	&0.021	&0.862	&0.863	&0.072	&0.898	&0.891	&0.045

% \\
% 8 &\checkmark &\checkmark  &\checkmark & &\textbf{0.886}	&\textbf{0.891}	&\textbf{0.050}	&\textbf{0.921}	&\textbf{0.906}	&\textbf{0.027}	&\textbf{0.905}	&0.903	&\textbf{0.043}	&\textbf{0.938}	&\textbf{0.929}	&\textbf{0.020}	&\textbf{0.866}	&\textbf{0.865}	&\textbf{0.069}	&\textbf{0.898}	&\textbf{0.892}	&\textbf{0.044}

%   \\

		\bottomrule
		\hline
	\end{tabular}
	\vspace{0pt}
\end{table*}

\begin{table*}[htbp]
	\centering
	\caption{Effectiveness of the recalibration process $\textbf{F}_{rec}$ in DHA. The number below ``$\textbf{F}_{rec}$'' in the table means the times of recalibration. Specifically, zero means that no recalibration is conducted.}\vspace{-0.3cm}
% 	\vspace{0cm}
	\label{tab:ablation2}
	\footnotesize
    %\scriptsize
    %\tiny
	\renewcommand{\arraystretch}{0.7}
	\renewcommand{\tabcolsep}{0.68mm}
	\begin{tabular}{c|c|cccc|cccc|cccc|cccc|cccc|cccc}
		\hline\toprule
		\multirow{2}{*}{\#} 
		&\multirow{2}{*}{$\textbf{F}_{rec}$}
		&\multicolumn{4}{c|}{\textbf{SIP~\cite{D3Net}}}  &\multicolumn{4}{c|}{\textbf{NLPR~\cite{NLPR}}}
		&\multicolumn{4}{c|}{\textbf{NJU2K~\cite{NJU2K}}}
		&\multicolumn{4}{c|}{\textbf{RGBD135~\cite{RGBD135}}}
		&\multicolumn{4}{c|}{\textbf{LFSD~\cite{LFSD}}}
		&\multicolumn{4}{c}{\textbf{STERE~\cite{STERE}}}\\ &
        &$S_{\alpha}$
	    &$F_{\beta}^{\rm max}$
	    & $E_{\xi}^{\rm max}$ 
		&$\mathcal{M}$  
		&$S_{\alpha}$
	    &$F_{\beta}^{\rm max}$
	    & $E_{\xi}^{\rm max}$ 
		&$\mathcal{M}$  
		&$S_{\alpha}$
	    &$F_{\beta}^{\rm max}$
	    & $E_{\xi}^{\rm max}$ 
		&$\mathcal{M}$  
		&$S_{\alpha}$
	    &$F_{\beta}^{\rm max}$
	    & $E_{\xi}^{\rm max}$ 
		&$\mathcal{M}$  
		&$S_{\alpha}$
	    &$F_{\beta}^{\rm max}$
	    & $E_{\xi}^{\rm max}$ 
		&$\mathcal{M}$  
		&$S_{\alpha}$
	    &$F_{\beta}^{\rm max}$
	    & $E_{\xi}^{\rm max}$ 
		&$\mathcal{M}$
	   
\\\midrule
5 &0&0.877	&0.881	&0.919	&0.054	&0.920	&0.904	&0.953	&0.027	&0.904	&0.904	&0.943	&0.043	&\textbf{0.934}	&\textbf{0.924}	&\textbf{0.973}	&0.021	&0.855	&0.855	&0.894	&0.076	&0.897	&0.890	&0.940	&0.045\\
6 &1&0.872	&0.873	&0.918	&0.056	&0.920	&0.904	&0.957	&0.027	&0.903	&0.902	&0.943	&0.043	&0.929	&0.922	&0.973	&0.023	&0.864	&\textbf{0.865}	&0.902	&0.072	&0.900	&0.893	&0.942	&0.044

\\
		\rowcolor{mygray}
		4 &2 &\textbf{{0.883}}	&\textbf{{0.887}}	&\textbf{{0.926}}	&\textbf{{0.051}}	&\textbf{{0.923}}	&\textbf{{0.908}}	&\textbf{0.957}	&\textbf{{0.026}}	&\textbf{{0.906}}  &\textbf{{0.910}}	&\textbf{{0.947}}	&\textbf{{0.042}}	&0.931	&0.922	&0.972	&\textbf{0.021}	&\textbf{{0.865}}	&0.864	&\textbf{{0.903}}	&\textbf{{0.072}}	&0.898	&0.893	&0.941	&0.045
\\
7 &3 &0.864 &0.868  &0.916 &0.059&0.921  &0.9068  &0.957 &0.026 &0.898  &0.897  &0.941  &0.042	 	   &0.917  &0.905  &0.963  &0.025   &0.858  &0.858  &0.896  &0.077  &\textbf{0.905}  &\textbf{0.897}  &\textbf{0.946}  &\textbf{0.042}
\\
		\bottomrule
		\hline
	\end{tabular}
	\vspace{0pt}
	\label{tab:recalibration}
\end{table*}

\begin{figure}
\centering
\begin{overpic}[abs,scale=0.55,unit=1mm]{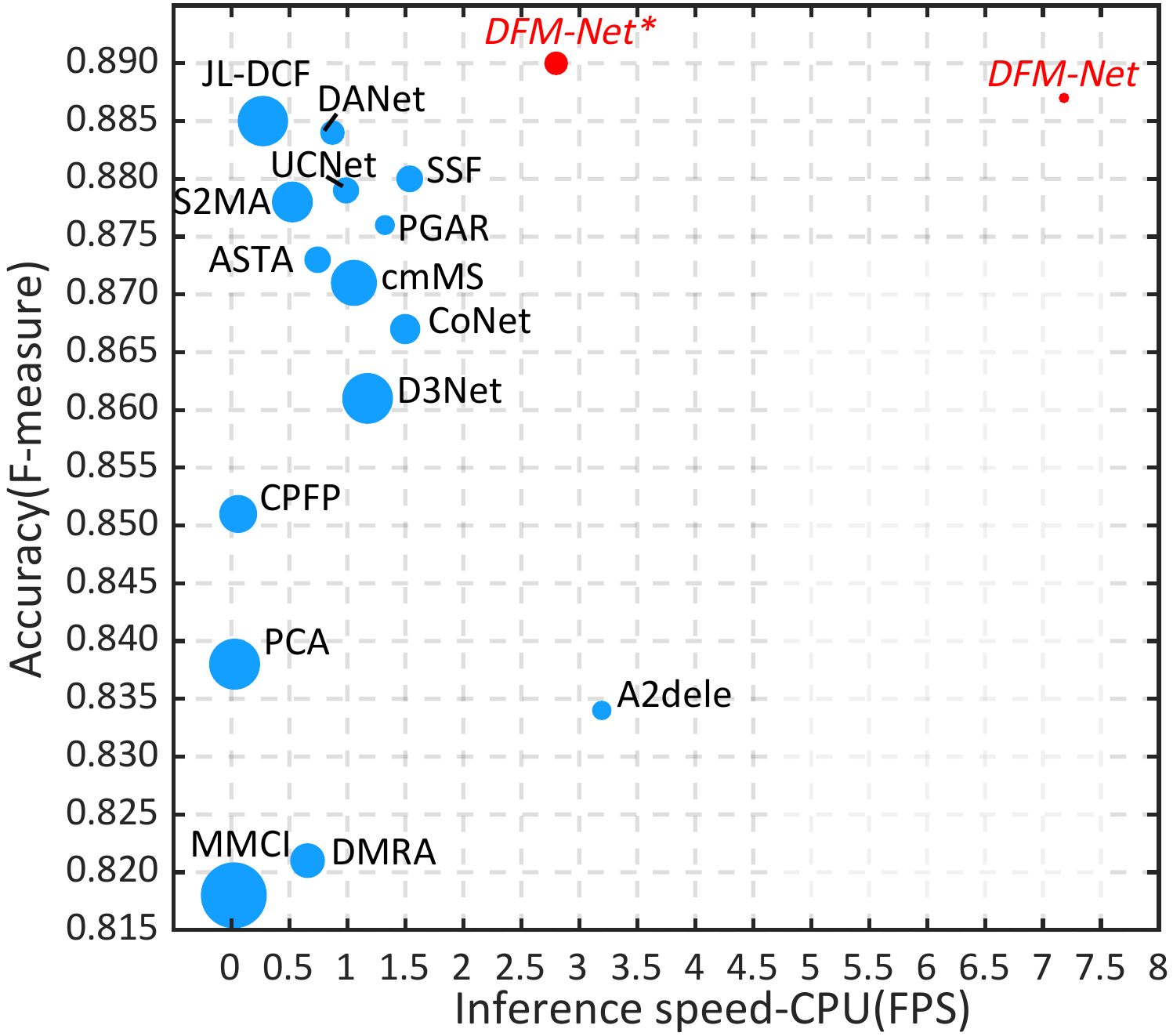}

\put(53,32){
	\footnotesize
    %\scriptsize
    %\tiny
	\renewcommand{\arraystretch}{0.7}
	\renewcommand{\tabcolsep}{0.5mm}
	\begin{tabular}{l|c}
		\hline\toprule
 Methods& Model size 
\\
		\midrule
MMCI~\cite{MMCI} &930\\PCA~\cite{PCF} &534\\D3Net~\cite{D3Net} &530\\JL-DCF~\cite{JLDCF} &520\\cmMS~\cite{cmMS} &430\\S2MA~\cite{S2MA} &330\\CPFP~\cite{CPFP} &278\\DMRA~\cite{DRMA} &228\\CoNet~\cite{CoNet} &167\\SSF~\cite{SSF} &125\\ASTA~\cite{ATSA} &123\\UCNet~\cite{UCNet} &119\\DANet~\cite{DANet} &102\\
\rowcolor{mygray}
\textit{\textbf{DFM-Net*}} &\textbf{93}\\\midrule PGAR~\cite{PGAR} &62\\A2dele~\cite{A2dele} &57\\
\rowcolor{mygray}
\textit{\textbf{DFM-Net}} &\textbf{8.5}\\
		\bottomrule
		\hline
	\end{tabular}}
\end{overpic}

\caption{Performance visualization. The vertical axis indicates the accuracy ($F_{\beta}^{\rm max}$) on SIP~\cite{D3Net}. The horizontal axis indicates the CPU speed (FPS). The circle area is proportional to the model size. More details refer to Tab.~\ref{tab:benchmark}. }\vspace{0cm}
\label{fig:benchmark}
\end{figure} 

\vspace{-5pt}
\subsection{Implementation Details}

Experiments are conducted on a workstation with Intel Core i7-8700 CPU, Nvidia GTX 1080Ti GPU with CUDA 10.1. We implement \ourmodel~by PyTorch \cite{pytorch}, and RGB and depth images are both resized to $256 \times 256$ for input. For testing, the inference time on CPU and FPS on GPU is obtained by averaging 100 times inference with batch size 1, and no any data augmentation or post processing is used. For training, in order generalize the network on limited training samples, following ~\cite{BBSNet}, we apply various data augmentation techniques \ie{}random translation/cropping, horizontal flipping, color enhancement and so on. We train \ourmodel~for 300 epochs on a single 1080Ti GPU, taking about 4 hours. The initial learning rate is set as 1e-4 for Adam optimizer, and the batch size is 10. The poly learning rate policy is used, where the power is set to 0.9.

\subsection{Quantitative and Qualitative Comparisons}\label{sec:comparison}

Since we cannot expect that an extremely light-weight model can always outperform existing non-efficient models (which are much larger), we have also extended \ourmodel~to obtain a larger but more powerful model called \ourmodel*, by replacing MobileNet-v2 used in the RGB branch with ResNet-34~\cite{ResNet}. Then to align the features from the RGB branch to those from TDB, slight modifications are made.  Results of \ourmodel~and \ourmodel*, compared to 15 state-of-the-art (SOTA) models including PCF \cite{PCF}, MMCI \cite{MMCI}, CPFP \cite{CPFP}, DRMA \cite{DRMA}, D3Net \cite{D3Net}, JL-DCF \cite{JLDCF}, UCNet \cite{UCNet}, SSF \cite{SSF}, S2MA \cite{S2MA}, CoNet \cite{CoNet}, cmMS \cite{cmMS}, DANet \cite{DANet}, ATSA \cite{ATSA}, A2dele \cite{A2dele} and PGAR \cite{PGAR}, can be found in Tab. \ref{tab:benchmark}.

As shown in Tab. \ref{tab:benchmark}, \ourmodel~surpasses existing efficient models A2dele \cite{A2dele} and PGAR \cite{PGAR} significantly on detection accuracy and model size, as well as CPU speed. It runs at 140ms on CPU, which is the fastest among all the contenders, with only $\sim$8.5Mb model size (14.9\% of that of A2dele). 
About GPU speed, \ourmodel~ranks the second after A2dele \cite{A2dele}. This is because the depth-wise separable convolution layers, which have been extensively used in MobileNet-v2 and TDB, cannot be fully accelerated on GPU in the current implementation \cite{Rooflinemodel,hong2021deep}. On the other hand, one can see that \ourmodel* achieves SOTA performance when compared to existing non-efficient models, with absolutely the fastest CPU/GPU speed and the smallest model size. 
Visual comparisons with representative methods are shown in Fig.~\ref{fig:visual_results}, where our results are closer to the ground truth (GT). 

To better reflect the advantages of the proposed method, as shown in Fig. \ref{fig:benchmark}, we visualize all models by plotting their accuracy ($F_{\beta}^{\rm max}$ on SIP dataset, the vertical axis), \emph{w.r.t.} their CPU speed (FPS is used here for better visualization, the horizontal axis) and model size (proportional to the circle diameter). It is clearly seen that \ourmodel~and \ourmodel* can rank the most upper right with very small circles, indicating that our method can perform well in terms of both efficiency and accuracy when compared to existing techniques.

\begin{table*}[htbp]
	\centering
	\caption{DQFM gating strategy: using identical (only one) $\alpha_i$ and $\beta_i$ \emph{vs.} using multiple (five different) $\alpha_i$ and $\beta_i$. }\vspace{-0.1cm}
	\label{tab:ablation3}
	\footnotesize
    %\scriptsize
    %\tiny
	%\renewcommand{\arraystretch}{}
	\renewcommand{\tabcolsep}{0.6mm}
	\begin{tabular}{c|c|cccc|cccc|cccc|cccc|cccc|cccc}
		\hline\toprule
		\multirow{2}{*}{\#} 
		&\multirow{2}{*}{Strategy}
		&\multicolumn{4}{c|}{\textbf{SIP~\cite{D3Net}}}  &\multicolumn{4}{c|}{\textbf{NLPR~\cite{NLPR}}}
		&\multicolumn{4}{c|}{\textbf{NJU2K~\cite{NJU2K}}}
		&\multicolumn{4}{c|}{\textbf{RGBD135~\cite{RGBD135}}}
		&\multicolumn{4}{c|}{\textbf{LFSD~\cite{LFSD}}}
		&\multicolumn{4}{c}{\textbf{STERE~\cite{STERE}}}\\ &
        &$S_{\alpha}$
	    &$F_{\beta}^{\rm max}$
	    & $E_{\xi}^{\rm max}$ 
		&$\mathcal{M}$  
		&$S_{\alpha}$
	    &$F_{\beta}^{\rm max}$
	    & $E_{\xi}^{\rm max}$ 
		&$\mathcal{M}$  
		&$S_{\alpha}$
	    &$F_{\beta}^{\rm max}$
	    & $E_{\xi}^{\rm max}$ 
		&$\mathcal{M}$  
		&$S_{\alpha}$
	    &$F_{\beta}^{\rm max}$
	    & $E_{\xi}^{\rm max}$ 
		&$\mathcal{M}$  
		&$S_{\alpha}$
	    &$F_{\beta}^{\rm max}$
	    & $E_{\xi}^{\rm max}$ 
		&$\mathcal{M}$  
		&$S_{\alpha}$
	    &$F_{\beta}^{\rm max}$
	    & $E_{\xi}^{\rm max}$ 
		&$\mathcal{M}$
	   
\\\midrule
		8 &Identical &0.880	&0.884	&0.924	&0.053	&0.922	&0.907	&\textbf{{0.958}}	&0.026	&0.901	&0.901	&0.944	&0.044	&0.924	&0.911	&0.961	&0.024	&0.859	&0.857	&0.899	&0.073	&0.898	&0.892	&\textbf{{0.942}}	&0.045
		
\\
		\rowcolor{mygray}
		4 &Multiple &\textbf{{0.883}}	&\textbf{{0.887}}	&\textbf{{0.926}}	&\textbf{{0.051}}	&\textbf{{0.923}}	&\textbf{{0.908}}	&0.957	&\textbf{{0.026}}	&\textbf{{0.906}}  &\textbf{{0.910}}	&\textbf{{0.947}}	&\textbf{{0.042}}	&\textbf{{0.931}}	&\textbf{{0.922}}	&\textbf{{0.972}}	&\textbf{{0.021}}	&\textbf{{0.865}}	&\textbf{{0.864}}	&\textbf{{0.903}}	&\textbf{{0.072}}	&\textbf{{0.898}}	&\textbf{{0.893}}	&0.941	&\textbf{{0.045}}
\\

		\bottomrule
		\hline
	\end{tabular}
	\vspace{0pt}
\end{table*}

\begin{table*}[htbp]
	\centering
	\caption{Performance of the proposed TDB (tailored depth backbone) against MobileNet-V2. Details are in Sec.~\ref{sec:ablation}.}\vspace{-0.1cm}
	\label{tab:ablation4}
	
	\footnotesize
    %\scriptsize
    %\tiny
	%\renewcommand{\arraystretch}{}
% 	\renewcommand{\arraystretch}{0.12}
	\renewcommand{\tabcolsep}{0.4mm}
	\begin{tabular}{c|c|c|cccc|cccc|cccc|cccc|cccc|cccc}
		\hline\toprule
		\multirow{2}{*}{\#} 
% 		&\multirow{2}{*}{Depth }
% 		&\multirow{2}{*}{Size} 		

        &Depth
		&Size

		&\multicolumn{4}{c|}{\textbf{SIP~\cite{D3Net}}}  &\multicolumn{4}{c|}{\textbf{NLPR~\cite{NLPR}}}
		&\multicolumn{4}{c|}{\textbf{NJU2K~\cite{NJU2K}}}
		&\multicolumn{4}{c|}{\textbf{RGBD135~\cite{RGBD135}}}
		&\multicolumn{4}{c|}{\textbf{LFSD~\cite{LFSD}}}
		&\multicolumn{4}{c}{\textbf{STERE~\cite{STERE}}}\\ &backbone &(Mb)
        &$S_{\alpha}$
	    &$F_{\beta}^{\rm max}$
	    & $E_{\xi}^{\rm max}$ 
		&$\mathcal{M}$  
		&$S_{\alpha}$
	    &$F_{\beta}^{\rm max}$
	    & $E_{\xi}^{\rm max}$ 
		&$\mathcal{M}$  
		&$S_{\alpha}$
	    &$F_{\beta}^{\rm max}$
	    & $E_{\xi}^{\rm max}$ 
		&$\mathcal{M}$  
		&$S_{\alpha}$
	    &$F_{\beta}^{\rm max}$
	    & $E_{\xi}^{\rm max}$ 
		&$\mathcal{M}$  
		&$S_{\alpha}$
	    &$F_{\beta}^{\rm max}$
	    & $E_{\xi}^{\rm max}$ 
		&$\mathcal{M}$  
		&$S_{\alpha}$
	    &$F_{\beta}^{\rm max}$
	    & $E_{\xi}^{\rm max}$ 
		&$\mathcal{M}$
	   
\\\midrule
		9 &MoblieNet-V2 &6.9 &0.879	&0.886	&0.923	&0.054	&0.919	&0.904	&0.954	&0.027	&0.906	&0.908	&\textbf{{0.948}}	&0.042	&0.930	&0.922	&0.969	&0.022	&0.864	&0.864	&0.902	&0.070	&0.893	&0.889	&\textbf{{0.942}}	&0.046
		
\\
		\rowcolor{mygray}
		4 &Tailored &0.9 &\textbf{{0.883}}	&\textbf{{0.887}}	&\textbf{{0.926}}	&\textbf{{0.051}}	&\textbf{{0.923}}	&\textbf{{0.908}}	&\textbf{{0.957}}	&\textbf{{0.026}}	&\textbf{{0.906}}  &\textbf{{0.910}}	&0.947	&\textbf{{0.042}}	&\textbf{{0.931}}	&\textbf{{0.922}}	&\textbf{{0.972}}	&\textbf{{0.021}}	&\textbf{{0.865}}	&\textbf{{0.864}}	&\textbf{{0.903}}	&\textbf{{0.072}}	&\textbf{{0.898}}	&\textbf{{0.893}}	&0.941	&\textbf{{0.045}}

\\

		\bottomrule
		\hline
	\end{tabular}
	\vspace{0pt}
\end{table*}

\begin{figure}
  \centering
 \centerline{\epsfig{figure=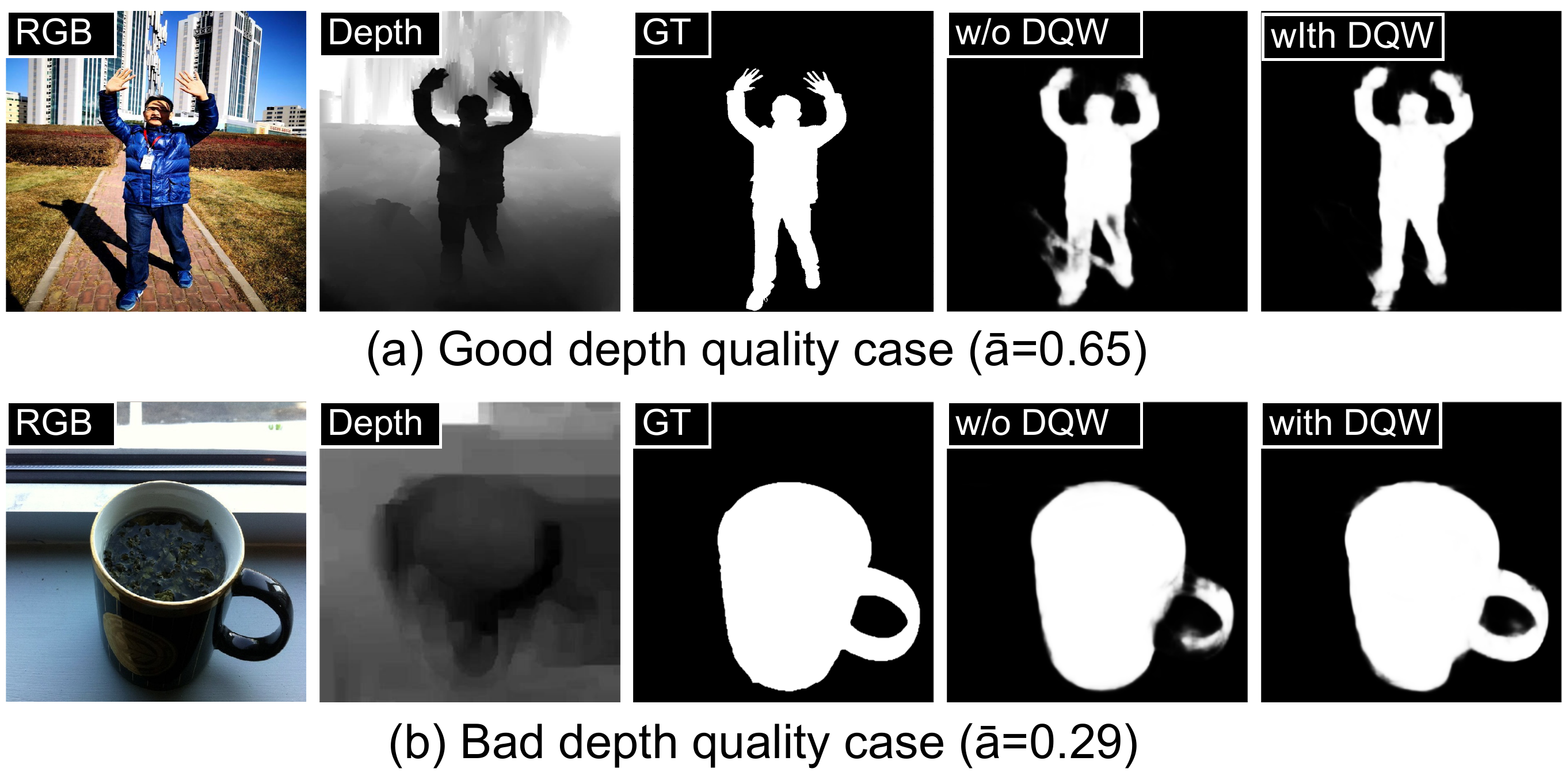,width=0.48\textwidth}}\vspace{-0.3cm}
\caption{Visualized examples of setting \#3 (w/o DQW) and \#4 (with DQW) for good (a) and bad (b) depth quality cases. $\overline{\alpha}$ denotes the average of $\alpha_1\sim \alpha_5$, which can be deemed as an indicator for depth quality.}\vspace{-0.2cm}
\label{fig:qualitative_dqw}
\end{figure} 

\vspace{-10pt}
\subsection{Ablation Studies}\label{sec:ablation}
We conduct thorough ablation studies on six datasets by removing or replacing components from the full implementation
of \ourmodel.

\textbf{Effectiveness of DQFM}. DQFM consists of two key components, namely DQW and DHA. Tab.~\ref{tab:ablation1} shows different configurations by ablating DQW/DHA. In detail, \#1 denotes a baseline model which has removed both DQW and DHA from \ourmodel. Configuration \#2 and \#3 mean having either one component, while \#4 means the full model of \ourmodel. 
Basically, from Tab.~\ref{tab:ablation1} one can see that incorporating either DQW and DHA into the baseline model \#1 leads to consistent improvement on almost all datasets. Besides, comparing \#2/\#3 to \#4, we see that employing both DQW and DHA can further enhance the results, demonstrating the complementary effect between DQW and DHA. The underlying reason could be that, although DHA is able to enhance potential target regions in the depth, it is unavoidable to make some mistakes (\emph{e.g.}, highlight wrong regions) especially in low depth quality cases. Luckily, DQW somewhat relieves such a side-effect because in the meantime it assigns low global weights to depth features. In all, these two components can work cooperatively to improve the robustness of the network, as we mentioned in Sec.~\ref{sec:DQFM}. Last but not least, from the model sizes of different configurations shown in Tab.~\ref{tab:ablation1} , the extra model parameters for introducing DQW and DHA are quite few. This rightly meets our goal and implies that they could potentially become universal components for light-weight models in the future.

In Fig.~\ref{fig:qualitative_dqw}, we show visual examples of setting \#3 (namely without DQW) and \#4. We also visualize the magnitudes of $\alpha$, namely the mean value $\overline{\alpha}$ of $\alpha_1 \sim \alpha_5$. From Fig.~\ref{fig:qualitative_dqw} (a) and (b), we can see that incorporating DQW does help improve detection accuracy, and practically, DQW is able to function as expected, namely rendering the good quality case with higher weights ($\overline{\alpha}=0.65$), and vice versa ($\overline{\alpha}=0.29$). In the good quality case (a), it is difficult to distinguish between the shadow and the man's legs in the RGB view, but this can be done easily in the depth view. Incorporating DQW to give more emphasis to depth features, therefore, can help better separate the entire human body from the shadow. In the bad quality case (b), although in the depth view the cup handle is much blurry, the impact of misleading depth has been alleviated, and the whole object still can be detected out accurately.  
%Comparing \#3 with \#4, we can see that DQW improves the accuracy greatly, which validates the effectiveness of DQW. To further illustrate its work mechanism, we display the depth quality case on both ends of the spectrum for setting \#3 and \#4 in Fig.~\ref{fig:qualitative_dqw}. Note that in the case (a), it is difficult to distinguish between the shadow and the man's leg in the RGB while there is little interfere of the shadow in the depth map of good quality. Thanks to DQW that can give a larger proportion to depth features when fused with RGB, \#4 better identifies the structure of the salient object in this challenging scene than \#3. On the other hand, in the case (b), depth is of bad quality, so DQW generate a lower $\alpha$ avoid influenced by noisy of depth. As a result, \#4 successfully detect the whole object from RGB while \#3 is disturbed by noisy depth. 
%Comparing \#2 with \#4, impressive improvement can be obtained when equipped with DHA, which validates the its effectiveness. We display the visual comparison in Fig.\ref{fig:qualitative_dha}, from the heat map of $\beta_1 \sim \beta_5$, we can see that \#4 with DHA can initially locate salient object\ie{}the girl, and filter out the messy background in the depth. Therefore, the final results avoid influence caused by the messy background.

\textbf{Recalibration in DHA}. As described in Sec.~\ref{sec:DHA}, in DHA, we utilize operation $\textbf{F}_{rec}$ to recalibrate the coarse information from high-level depth features. To validate the necessity of using $\textbf{F}_{rec}$, we experiment with different times (from 0 to 3) of using $\textbf{F}_{rec}$. These variants are denoted as \#5, \#6, \#4, and \#7 in  Tab.~\ref{tab:recalibration}. Note that \#4 corresponds to the default implementation of \ourmodel. From Tab. \ref{tab:recalibration}, 
we can see that \#4 (recalibrate twice) achieves the overall best performance. The underlying reason should be that, appropriate usage of $\textbf{F}_{rec}$ can expand the coverage areas of attention maps to make conservative filtering for object edges as well as some inaccurately located objects, but too large receptive field leads to over-dilated attention regions that are less informative. This is the reason why when the times increase to 3, the performance starts to degenerate on most datasets, except on STERE whose depth quality is generally low, which easily leads to inaccurate attention location.

\textbf{DQFM Gating Strategy}. As we mentioned in Sec.~\ref{sec:DQFM}, we  adopt a multi-variable strategy for $\alpha_i$ and $\beta_i$. To validate this strategy, we compare it to the single-variable strategy, namely using identical (only one) $\alpha_i$ and $\beta_i$. Tab.~\ref{tab:ablation3} shows the results, from which it can be seen that our proposed multi-variable strategy is better, because it somewhat increases the network flexibility by rendering different hierarchies with different quality weights and attention maps.

\textbf{Tailored Depth Backbone}. The effectiveness of TDB is validated by comparing it to MobileNet-V2. We implement a configuration \#9 by switching TDB directly to MobileNet-V2, while maintaining all other settings unchanged. Evaluation results are shown in Tab. \ref{tab:ablation4}. We can see that our tailored depth backbone is more efficient (0.9Mb vs. MobileNet-V2's 6.9Mb) and also more accurate (with noticeable gains) than MobileNet-V2 when utilized in \ourmodel. This demonstrates the feasibility of using a lighter backbone to process depth data for efficiency purpose.%as mentioned in Sec. \ref{sec:DepthBackbone}.

\section{Conclusion}
In this paper, we propose an efficient RGB-D SOD model called \ourmodel, characterized by the the DQFM process to explicitly control and enhance depth features during cross-modal fusion. The two key components in DQFM, namely DQW and DHA, are validated by comprehensive ablation experiments. The experimental results show that DQW and DHA are both essential for obtaining higher detection accuracy with very few model parameters added on. Besides, a tailored depth backbone and a two-stage decoder are elaborately designed to further improve the efficiency of \ourmodel. Our \ourmodel~achieves new state-of-the-art records on light-weight model size as well as CPU speed, meanwhile retaining decent accuracy. In the future, it is very attractive to apply \ourmodel~on some embedding or mobile systems that process RGB-D data.

\balance

\bibliographystyle{ACM-Reference-Format}
\bibliography{sample-base}

\end{document}